\documentclass[letterpaper,journal]{IEEEtran}
\usepackage{amsmath,amsfonts}
\usepackage{algorithmic}
\usepackage{algorithm}
\usepackage{array}
\usepackage[caption=false,font=normalsize,labelfont=sf,textfont=sf]{subfig}
\usepackage{textcomp}
\usepackage{stfloats}
\usepackage{url}
\usepackage{verbatim}
\usepackage{graphicx}
\usepackage{cite}
\usepackage{booktabs}
\usepackage{multirow}
\usepackage{threeparttable}

\usepackage[colorlinks,linkcolor=blue,citecolor=blue]{hyperref}

\usepackage[table,xcdraw]{xcolor}

\usepackage{dashrule}
\newcommand{\dashedline}{\hdashrule{\linewidth}{0.4pt}{3pt}}

\begin{document}
\title{DGFusion: Dual-guided Fusion for Robust Multi-Modal 3D Object Detection}

\author{Feiyang Jia, Caiyan Jia, Ailin Liu, Shaoqing Xu, Qiming Xia, \\Lin Liu, Lei Yang, Yan Gong, Ziying Song 
\thanks{\textbf{Notice: This is the author's version of a manuscript that has been accepted for publication in \emph{IEEE Transactions on Circuits and Systems for Video Technology}. 
The final version of record is available at \href{https://doi.org/10.1109/TCSVT.2025.3628019}{https://doi.org/10.1109/TCSVT.2025.3628019}. $\copyright$~2025 IEEE. For any other use, permission must be obtained from IEEE.}
}
\thanks{This work was supported in part by the National Key R\&D Program of China (2018AAA0100302). \emph{(Corresponding author: Caiyan Jia and Ziying Song.)}}
\thanks{Feiyang Jia, Caiyan Jia, Ailin Liu,  Lin Liu and Ziying Song are with School of Computer Science and Technology, Beijing Key Laboratory of Traffic Data Mining and Embodied Intelligence, Beijing Jiaotong University, Beijing 100044, China (e-mail: feiyangjia@bjtu.edu.cn; cyjia@bjtu.edu.cn; 24125249@bjtu.edu.cn;  liulin010811@gmail.com, 22110110@bjtu.edu.cn.).}
\thanks{Shaoqing Xu is with the State Key Laboratory of Internet of Things for Smart City and Department of Electrome chanical Engineering, University of Macau, Macau 999078, China (e-mail: shaoqing.xu@connect.um.edu.mo)}
\thanks{Qiming Xia is with Fujian Key Laboratory of Sensing and Computing for Smart Cities, Xiamen University, Xiamen, China, Fujian 361005, China (e-mail: qimingxia96@163.com).}
\thanks{Lei Yang is with School of Mechanical and Aerospace Engineering, Nanyang Technological University, Singapore. (email: lei.yang@ntu.edu.sg)}
\thanks{Yan Gong are with the State Key Laboratory of Robotics and System, Harbin Institute of Technology, Harbin 150001, China (email: gongyan2020@foxmail.com).}
}



\maketitle

\begin{abstract}
As a critical task in autonomous driving perception systems, 3D object detection is used to identify and track key objects, such as vehicles and pedestrians. However, detecting distant, small, or occluded objects (hard instances) remains a challenge, which directly compromises the safety of autonomous driving systems. We observe that existing multi-modal 3D object detection methods often follow a single-guided paradigm, failing to account for the differences in information density of hard instances between modalities. In this work, we propose DGFusion, based on the Dual-guided paradigm, which fully inherits the advantages of the Point-guide-Image paradigm and integrates the Image-guide-Point paradigm to address the limitations of the single paradigms. 
The core of DGFusion, the Difficulty-aware Instance Pair Matcher (DIPM), performs instance-level feature matching based on difficulty to generate easy and hard instance pairs, while the Dual-guided Modules exploit the advantages of both pair types to enable effective multi-modal feature fusion.
Experimental results demonstrate that our DGFusion outperforms the baseline methods, with respective improvements of +1.0\% mAP, +0.8\% NDS, and +1.3\% average recall on nuScenes. Extensive experiments demonstrate consistent robustness gains for hard instance detection across ego-distance, size, visibility, and small-scale training scenarios.
\end{abstract}


\section{Introduction}
\begin{figure}[t]
\centering
\includegraphics[width=0.485\textwidth]{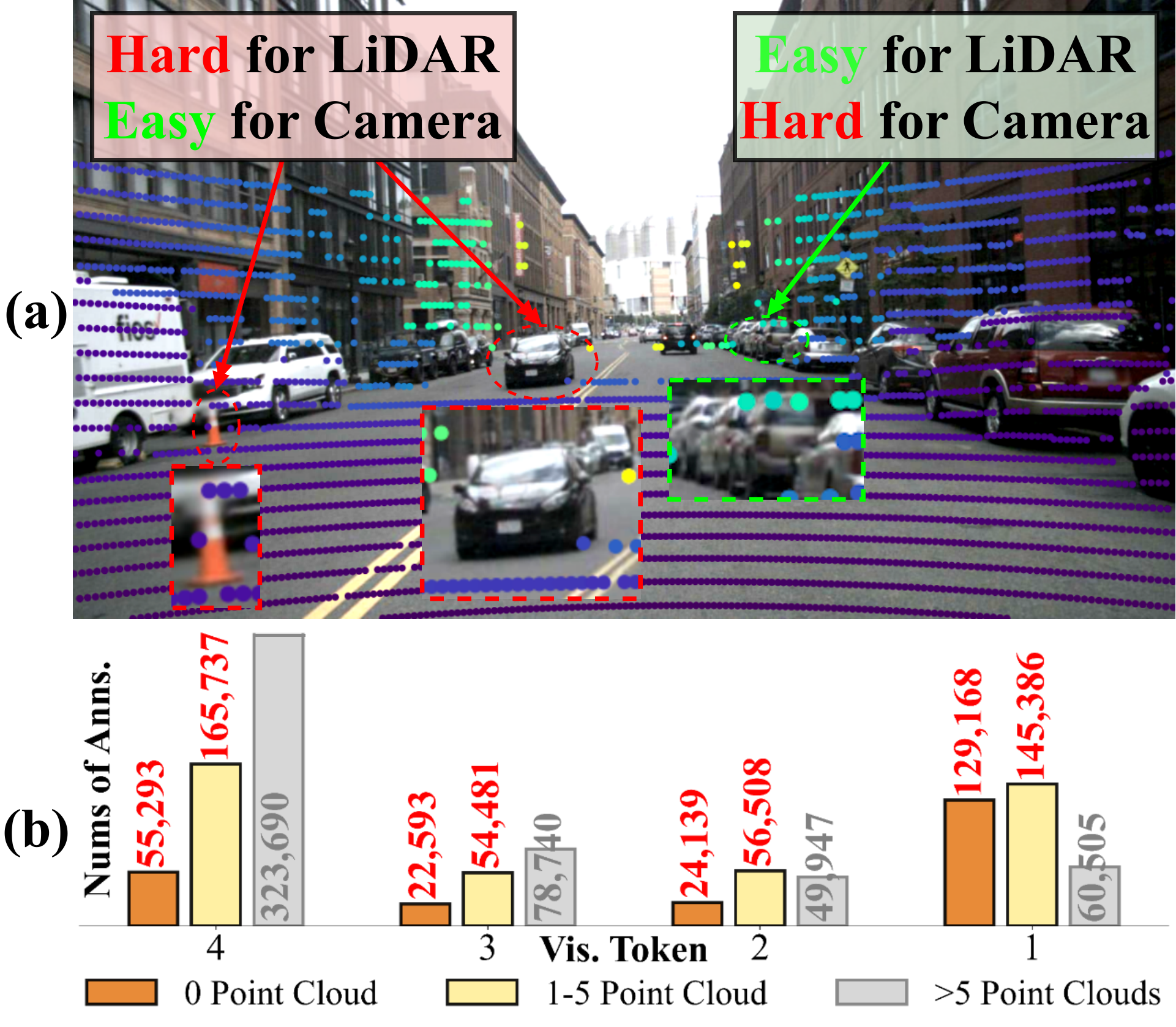}
\label{Sec:Introduction}
\caption[]{(a) The information density gap is a distinctive characteristic of certain distant, occluded, or small-scale targets. This phenomenon manifests as either poor point cloud data but rich pixel information (red dashed circle) or the converse scenario (green dashed circle). Most existing research focuses on a single case. 
(b) The number of point clouds from all annotations in the nuScenes training and validation sets is counted using the visibility tokens as a classification benchmark to demonstrate the generalization of the two phenomena mentioned above. Notably, even among objects with the highest visibility (token=4), over 20\% exhibits either zero or merely one LiDAR point. Conversely, a significant portion of objects with the lowest visibility (token=1) still retain rich point cloud data. Statement: 1) Picture from nuScenes\cite{nuscenes}, sample token: a771effa2a2648d78096c3e92b95b129, visualization and data statistics were implemented via Python SDK nuScenes DevKit\cite{nuscenes}. 2) For the key frames of the nuScenes LiDAR point clouds, the number of points falling within the bounding boxes of GT (ground-truth ) annotations is recorded under the attribute name `num\_lidar\_pts' - the value we count. 3) The visibility token, an attribute within the nuScenes annotations, quantifies the visibility level of targets in camera data, categorized as follows: 1 (0\%–40\%), 2 (40\%–60\%), 3 (60\%–80\%), and 4 (80\%–100\%).
}
\label{fig:idd}
\end{figure}

\begin{figure*}[t]
\centering
\includegraphics[width=0.85\textwidth]{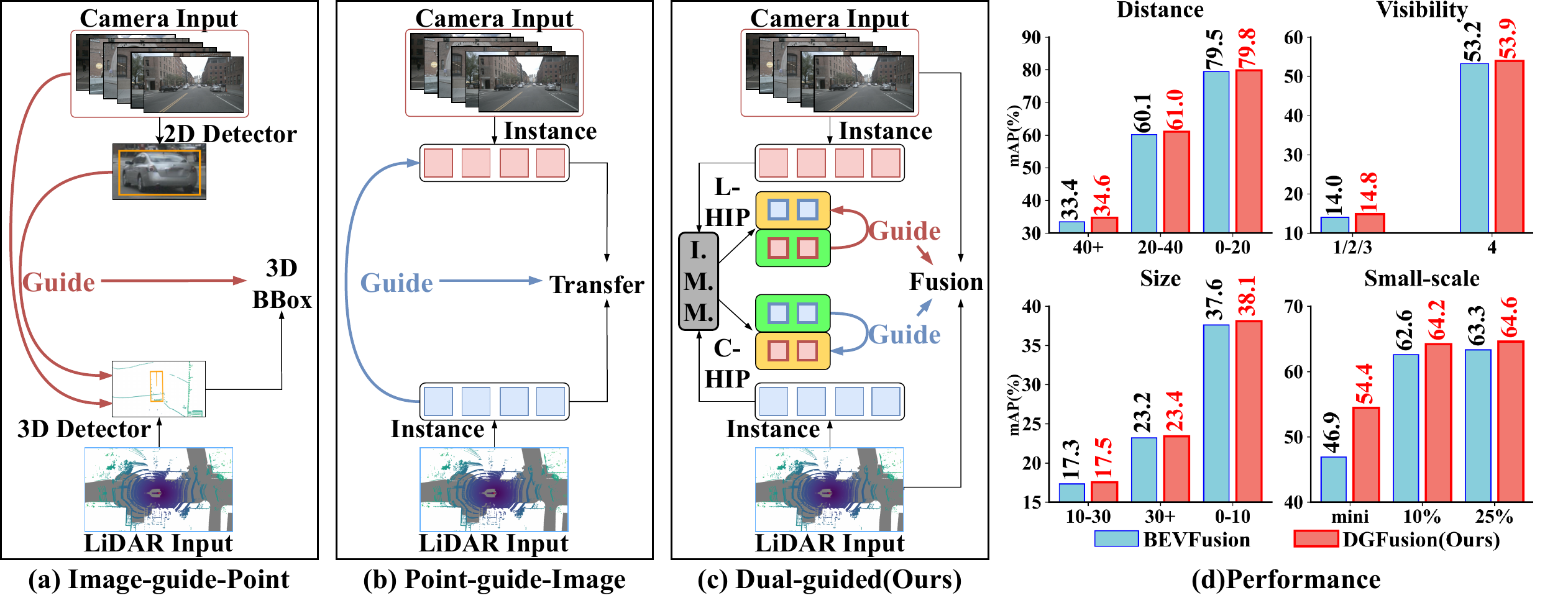}
\caption[]{
The paradigms of multi-modal 3D object detection methods in autonomous driving and the performance of our new paradigm.
(a) The Image-guide-Point paradigm obtains 2D feature information by human-designed elements to guide 3D feature information. (b) The Point-guide-Image paradigm acquires point cloud-dominated instance-level features in a BEV unified space to transfer semantic and geometric information. (c) The Dual-guided paradigm we propose can sensitively capture the information density gap between different modalities. (d) The \textbf{DGFusion} designed based on the Dual-guided paradigm, demonstrates exceptional robustness, without requiring additional training epochs.  Specifically, DGFusion's inference results on objects of varying distance (top left), visibility (top right), and size (bottom left) validate its effectiveness in mitigating hard instance detection challenges. Furthermore, all models generated by DGFusion using the nuScenes small-scale training dataset consistently exhibit stable and superior performance on the validation set (bottom right).} 
\label{fig:motivation}
\end{figure*}


3D object detection is a crucial component of perception tasks in autonomous driving. 
It is used to identify and track essential objects such as vehicles and pedestrians. The detection results help decision-making and control systems to ensure vehicle safety\cite{song2024robustness,mao20233d_zongshu,qian20223d}. 
Currently, the methods, based on single modality detection\cite{CBGS,Pointpillars,liu2020smoke,song2023vp,yang2023lite,yang2023mixteaching,chen2023focalformer3d,liu2024sparsedet,hou2025polarbevu,jiang2024far3d,ji2024enhancing,Centerpoint,wang2025rethinking,fan2023hcpvf,sheng2023pdr} and heterogeneous modality fusion\cite{mvp,ma2023long_LT3D_2dlatefusion,peri2023towardsLT3D,bevfusion-mit,yang2023bevheight,bevfusion-pku,bai2022transfusion,graphalign,graphalign++,song2024robofusion,wang2024mv2dfusion,UniTR,ObjectFusion,yin2024isfusion,qi2018frustum_pointnets,wang2020high_7Dpointnet,pointpainting,wang2021pointaugmenting,liu2022pai3d,cao2024tfienet,yang2024ralibev,huang2025l4dr,gu2025hgsfusion}, have demonstrated outstanding performance on benchmark datasets, including KITTI\cite{kitti}, nuScenes\cite{nuscenes}, and Waymo\cite{Waymo}. 
Although multi-modal 3D object detection has made significant progress recently by taking advantage of structured data collected by sensors such as cameras and LiDAR, 
efforts to address hard instance detection have not been paid enough attention, which are of critical importance and directly related to the safety of autonomous driving. 
As evidenced by \cite{chen2023focalformer3d}, hard instances significantly increase the rate of missing detection in perception systems, which means that decision-making and control systems cannot acquire information about potential obstacles, thus increasing the probability of collisions. Yet, hard instance detection caused by long-range, small, or occluded objects, remains a significant challenge. 

An ideal solution to advance hard instance detection is multi-modal 3D object detection based on LiDAR point clouds and camera images \cite{ma2023long_LT3D_2dlatefusion}. This is because multi-modal object detection incorporates information from point clouds and images, where point clouds provide valuable depth information\cite{lift_splat_shoot} for precise localization and images offer complementary information for object recognition, disambiguating geometrically similar but semantically different objects in point clouds. 
A pivotal characteristic of challenging instances is the pronounced information density gap across different modalities, as illustrated in Fig.\ref{fig:idd}.  
For example, the raw information in point clouds for small distant objects is much sparser than the images of the same objects (see the traffic cone in Fig.\ref{fig:idd}.(a)). Sometimes, an information hole may occur in point clouds, e.g., the car in the middle of Fig.\ref{fig:idd}.(a), but as can be seen, the car is very clear in pixels. In these two cases, the detection tasks are hard for LiDAR, but easy for camera. In contrast, for occluded objects (e.g., the cars in the right part of Fig.\ref{fig:idd}.(a)), they are easy to be detected in LiDAR modality, but very hard for camera. The two distinct manifestations of the information density gap are prevalent (see Fig.\ref{fig:idd}.(b)). 

In the literature, there are two different paradigms to mitigate the gap, thereby achieving performance improvement. One is `Image-guide-Point' paradigm which uses image information to dominate the fusion process. The other is `Point-guide-Image' paradigm which takes the LiDAR branch as the main information source. 
Following the `Image-guide-Point' paradigm, some studies used 2D feature information to guide 3D feature information to achieve hard instance detection \cite{ma2023long_LT3D_2dlatefusion,peri2023towardsLT3D,gupta2023far3det}, as shown in Fig.\ref{fig:motivation}.(a). This paradigm emphasizes the inherent advantages of image data and has achieved initial success in hard instance detection. Yet, these methods do not integrate multi-modal data into a unified feature space, but instead, rely on human-designed elements such as 2D detectors \cite{ma2023long_LT3D_2dlatefusion,peri2023towardsLT3D} or prior knowledge \cite{gupta2023far3det}, which limit the model's generalization ability. Other studies \cite{sparsefusion,ObjectFusion,yin2024isfusion,song2024contrastalign} first construct unified representations of multi-modal data based on bird's eye view (BEV) and then transfer semantic and geometric information across modalities by building instance-level features, as shown in Fig.\ref{fig:motivation}.(b). During the fusion phase, the distinctive characteristics of hard instances originate primarily from point clouds. Therefore, we regard these methods as following the `Point-guide-Image' paradigm. Either of the two paradigms does not fully consider the significant impact of information density gap between modalities on hard instance detection. In other words, the aforementioned efforts have predominantly focused on only one of the phenomena in Fig.\ref{fig:idd} (a), neglecting a comprehensive exploration of both.

Based on the above discussion, fully considering the gap in information density between modalities might be the optimal approach to improving hard instance detection performance. 
Therefore, we propose DGFusion, a multi-modal 3D object detection framework that adheres to the Dual-guided paradigm, as illustrated in Fig.\ref{fig:motivation}.(c). 
Specifically, we construct instance-level features from multi-modal BEV features through an Instance-level Features Generator (IFG). Then we design a Difficulty-aware Instance Pair Matcher (DIPM) to match the instance-level features, forming easy instance pairs (EIP) and two types of hard instance pairs (HIP) based on different guiding paths, while eliminating redundant instances. Prior to BEV feature fusion, EIP/C-HIP (Camera-centric HIP) and L-HIP (LiDAR-centric HIP) are used for Point-guide-Image Enhancement (PGIE), and Image-guide-Point Enhancement (IGPE), respectively, to improve hard instance detection without compromising regular performance. The empirical studies on nuScenes have validated the superior performance of our paradigm.

Our contributions are summarized as follows.  
\begin{itemize}
\item 
We summarize the paradigms of existing methods and propose a novel paradigm, termed the Dual-guided paradigm. This paradigm aims to bridge the information density gap between modalities to enhance robustness.
\item
Based on the Dual-guided paradigm, we introduce DGFusion, a multi-modal 3D object detection framework, which effectively learns from multi-modal data and enhances BEV features efficiently. 
\item
Our evaluation of DGFusion on the nuScenes benchmark reveals significant performance gains, with improvements of 1.0\%, 0.8\%, and 1.3\% in mAP, NDS, and average recall, respectively. Furthermore, the framework exhibits strong robustness, as demonstrated in Fig.\ref{fig:motivation}.(d), compared with our baseline BEVFusion\cite{bevfusion-mit}.
\end{itemize}

\section{Related work}

\subsection{3D Object Detection in Autonomous Driving Perception}



Early autonomous driving perception algorithms tend to rely on LiDAR-based detectors\cite{Pointrcnn,Pointpillars,deng2021multi_H23DRCNN,RT3D,Voxelnet,wang2023auto-points}. Some methods\cite{RT3D,Voxelnet,Cia-ssd,SSN} employ anchor-based model architectures, which define anchor boxes of various shapes to guide class-specific object detection. Anchor-free models \cite{Centerpoint,afdet,AFDetV2,fan2023hcpvf,chen2020objecthotspots,deng2021multi_H23DRCNN} eliminate the manual design of anchor boxes, simplifying the training process and improving the adaptability of the model to objects of different scales. For example, CenterPoint \cite{Centerpoint} locates objects by learning to predict the centers of the targets. HCPVF\cite{fan2023hcpvf} adopts a hierarchical cascaded point-voxel fusion module with a two-layer detection head to balance accuracy and speed. $H^2$3D RCNN\cite{deng2021multi_H23DRCNN} captures long-range dependencies between features in 2D feature space to improve efficiency. Although LiDAR sensors provide depth information, they cannot directly capture complex geometric and ego-motion cues\cite{peri2023towardsLT3D,khan2024lidar}.

Camera sensors do not directly provide precise depth information, which makes early Camera-based detectors \cite{liu2020smoke,wang2021fcos3d,CaDDN,zhang2021objects} struggle with accurately estimating the distance and shape of objects, leading to overall performance that is generally inferior to LiDAR-based detectors. However, camera detectors excel in detecting rare object categories\cite{ma2023long_LT3D_2dlatefusion,peri2023towardsLT3D,kim2025labeldistill}, and recent camera-based methods \cite{ji2024enhancing,yang2023lite,park2024odd,li2024unimode,brazil2023omni3d,FSD,FSDV2,yang2023mixteaching,sheng2023pdr,huang2021b,jiang2024far3d} have shown significant improvements in handling complex traffic scenarios. For example, PDR\cite{sheng2023pdr} achieves more accurate depth estimation through progressive regularization.

Multi-modal 3D object detection enhances object detection by leveraging and integrating the data features of heterogeneous sensors. Recently, BEV-based (Bird's Eye View) methods \cite{bevdepth,xu2021fusionpainting,xu2024multi,bevfusion-mit,bevfusion-pku,song2024graphbev,song2024robofusion,graphalign,graphalign++} have effectively merged LiDAR and Camera representations into BEV space, achieving the state-of-the-art (SOTA) performance. Although pioneers such as BEVFusion\cite{bevfusion-mit} have demonstrated high performance, typically evaluated on datasets like nuScenes, they overlook real-world complexities, particularly the issue of `hard instance detection', which presents a challenge for practical deployment. 

\subsection{Guided Paradigm}
Dividing existing work into the Image-guide-Point paradigm and the Point-guide-Image paradigm serves as the theoretical foundation of our DGFusion. LiDAR data inherently possesses reliable depth information; therefore, the performance of single-modal models that take point clouds as input is higher than that of models using images as input. This phenomenon has become a consensus among most researchers\cite{song2024robustness}. However, we have realized that when these two modalities are fused in a unified space (such as the widely used BEV paradigm), the same feature processing method conversely leads to an imbalance between the modalities. Meanwhile, most detection models with a unified feature space usually do not take into account the imbalance between features. The aforementioned studies are categorized by us into the Point-guide-Image paradigm\cite{bevfusion-mit,bevfusion-pku,ObjectFusion,chen2023focalformer3d,graphalign,graphalign++,bai2022transfusion,cao2024tfienet,UniTR}.

A small number of researchers further point out that in the fusion paradigm, image data is more conducive to hard instance detection, and we categorize these studies into the Image-guide-Point paradigm\cite{mvp,pointpainting,qi2018frustum_pointnets,wang2020high_7Dpointnet,wang2021pointaugmenting,peri2023towardsLT3D,ma2023long_LT3D_2dlatefusion}. Early works typically adhere to input-level fusion \cite{mvp,pointpainting,qi2018frustum_pointnets,wang2020high_7Dpointnet}. MVP\cite{mvp} is a core work in input-level fusion, whose design idea is to convert 2D detection results into 3D virtual point clouds to enhance the original point clouds. Recent works have shown that feature-level fusion \cite{wang2021pointaugmenting,peri2023towardsLT3D,ma2023long_LT3D_2dlatefusion} is more effective than input-level fusion.

Emerging as a novel third paradigm, our proposed dual-guided paradigm in this study advances beyond existing fusion strategies by establishing bidirectional feature interaction between point clouds and images.

\subsection{Hard Instance Detection}


The challenge of hard instance detection in autonomous driving scenarios has driven extensive research across multiple paradigms, as evidenced by recent advances in sensor fusion and deep learning architectures\cite{jiang2024far3d,gupta2023far3det,liu2024sparsedet,chen2023focalformer3d,song2024robofusion,graphalign++,UniTR,peri2023towardsLT3D,ma2023long_LT3D_2dlatefusion,peri2023empirical,liu2023generalized,pan2024clipbevformer}. Current solutions predominantly bifurcate into single-modality approaches and multi-modal approaches, each demonstrating distinct advantages in addressing specific aspects of this complex problem.

Within the single-modality domain, camera-based and LiDAR-centric methodologies have achieved notable progress. Far3D\cite{jiang2024far3d} establishes new benchmarks for long-distance detection through its camera-only implementation, particularly excelling on the Argoverse 2 dataset. Mix-Teaching\cite{yang2023mixteaching} can outperform the baseline performance with only 10\% of KITTI\cite{kitti} monocular data for training. Its principle is to merge pseudo-labels from multiple frames into a single image for the semi-supervised training of the student model, and select reliable pseudo-labels through an uncertainty-based filter. Complementarily, LiDAR-oriented approaches like FocalFormer3D\cite{chen2023focalformer3d} introduce systematic solutions through Hard Instance Probing (HIP), a multi-stage pipeline, which employs false negative identification to optimize detection performance. LDFCNet\cite{wang2025rethinking} transfers the complex task of capturing long-range dependencies to 2D dense feature maps for processing, significantly reducing the computational cost. While these single-modality strategies demonstrate modality-specific efficacy, their inherent sensor limitations naturally motivate the exploration of multi-modal alternatives.

Transitioning to multi-modal paradigms, current research primarily evolves along two methodological axes. The predominant Point-guide-Image paradigm emphasizes LiDAR feature dominance in fusion architectures, with representative works demonstrating enhanced spatial alignment capabilities. GraphAlign++\cite{graphalign++} systematically addresses long-range detection challenges through precision feature alignment, while UniTR\cite{UniTR} establishes sensor-agnostic robustness via unified 2D-3D relationship modeling. Contrastingly, the Image-guide-Point paradigm remains less explored despite its potential in addressing data scarcity issues. Ma {\it et al.}\cite{ma2023long_LT3D_2dlatefusion} reveal through comparative analysis that 2D RGB detectors surpass their 3D counterparts in rare category recognition, a finding corroborated by LT3D's\cite{peri2023towardsLT3D} hierarchical evaluation framework that introduces Hierarchical mAP to accommodate permissible classification errors. HGSFusion\cite{gu2025hgsfusion} aims to resolve the errors between the two modalities during camera-radar fusion. In fact, there are also some robustness works that follow the `Point-guide-Point' paradigm. RaLiBEV\cite{yang2024ralibev} alleviates the environmental perception problem under severe weather by fusing the features of Radar range-azimuth heatmap and LiDAR point cloud. L4DR\cite{huang2025l4dr} achieves weather robustness by fusing LiDAR and 4D radar. In general, these pioneering works demonstrate the necessity of the fusion paradigm in hard instance detection.

\section{Method}
\label{sec:method}
DGFusion, the robust fusion framework, focuses on addressing the issue of hard instance detection in 3D object detection. In the following sections, we first provide the framework of DGFusion in Sec.\ref{subsec:overallframework}. Subsequently, we elaborate on the design and implementation of the Instance Match Models in Sec.\ref{subsec:InstanceMatch}. Next, we conduct an in-depth analysis of the core design of Dual-guided paradigm in Sec.\ref{subsec:dg}. Finally, Sec.\ref{subsec:head} discloses the implementation details of the detection head and loss function.

\subsection{Overall Framework of DGFusion}
\label{subsec:overallframework}

\begin{figure*}[t]
\centering
\includegraphics[width=1\textwidth]{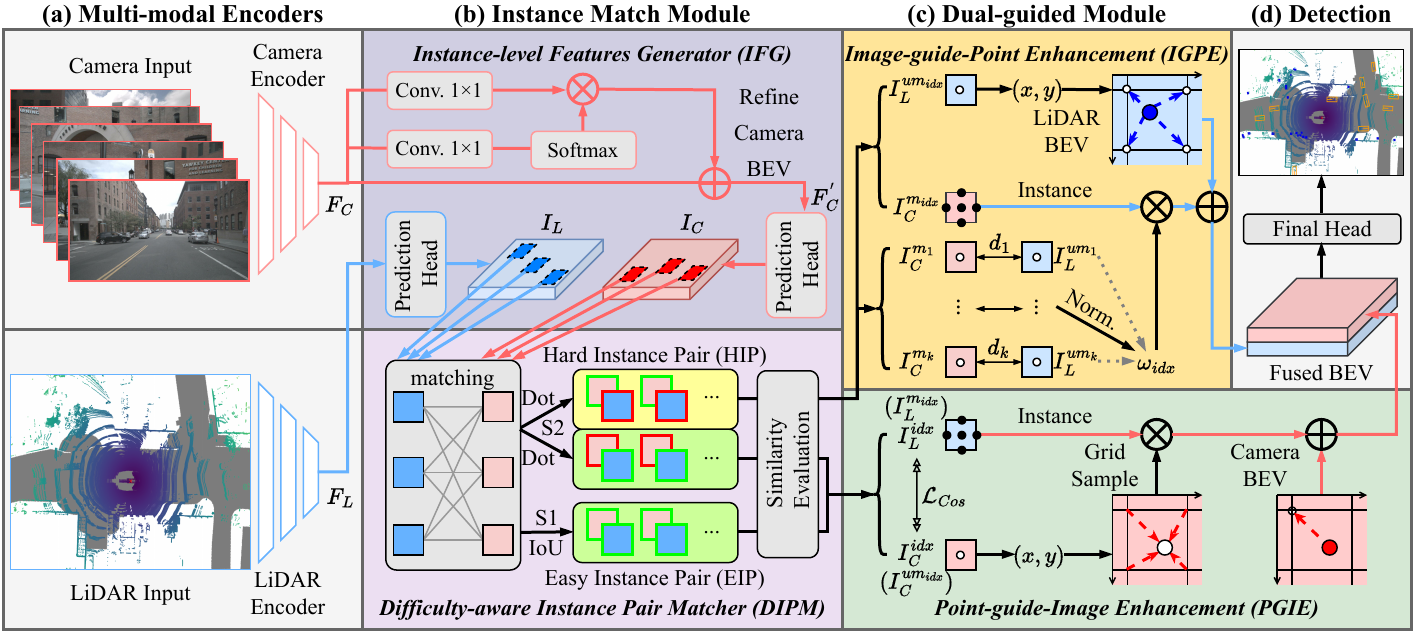}
\caption[]{\textbf{DGFusion Framework}. \textbf{(a)} First, we extract BEV features by structured integration of LiDAR and camera data. \textbf{(b)} The Instance Match Modules contains: (i) Instance-level Features Generator (IFG) that produces multi-modal instances, and (ii) Difficulty-aware Instance Pair Matcher (DIPM) that matches Easy Instance Pairs (EIP) and two Hard Instance Pairs types (C-HIP and L-HIP). \textbf{(c)} The Dual-guided Modules then performs: (i) Point-guide-Image Enhancement (PGIE) to enhance Camera BEV space using EIP and C-HIP, and (ii) Image-guide-Point Enhancement (IGPE) to enhance Camera BEV space using L-HIP. \textbf{(d)} Finally, we concatenate the enhanced BEV features and generate 3D detection results with a dense detection head.} 
\label{fig:framework}
\end{figure*}

\begin{figure}[t]
\centering
\includegraphics[width=0.48\textwidth]{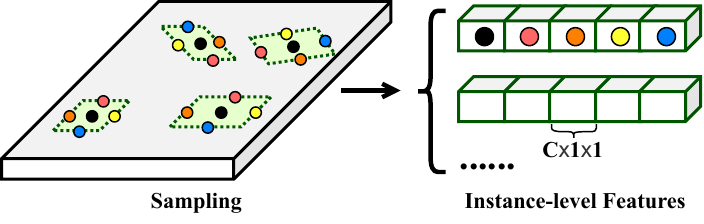}
\caption[ ]{
The process of \textbf{generating instance-level features} through sampling involves generating proposals directly from the BEV feature maps through the additional prediction head, project the obtained proposals onto the 2D space, acquire the center points of the proposals and the midpoints of each boundary line, and concatenate the features of these five key sample features to form enriched instance features.} 
\label{fig:CameraBEV_Enhancement_and_Instance_Generation}
\end{figure}

The overall framework, as shown in Fig.\ref{fig:framework}, is mainly composed of four modules: Multi-modal Encoders, Instance Match Modules, Dual-guided Modules, and Detection Head.

\textbf{Multi-modal Encoders.}
Multi-modal Encoders adhere to the broad definition to acquire Camera BEV features\cite{lift_splat_shoot} and LiDAR BEV features\cite{bai2022transfusion}, which are respectively defined as $F_{C} \in \mathbb{R}^{C_{C}\times H\times W}$ and $F_{L} \in \mathbb{R}^{C_{L}\times H\times W}$, where $H$ and $W$ denote the dimensions of the feature map, and $C_{C}$ and $C_{L}$ indicate the number of channels for Camera BEV features and LiDAR BEV features, respectively.

\textbf{Instance Match Modules.}
We aim to mitigate the hard instance detection problem by exploring the differences between easy instances and hard instances. The Instance Match Modules first utilizes BEV features to predict 3D proposals and then constructs instance features through a sampling operation and score filtering. Subsequently, we perform two-stage matching centered on LiDAR instances and camera instances respectively, to obtain EIP and two types of HIP for guided enhancement of BEV features. The Instance Match Modules represents our core innovation, with details provided in Sec.\ref{subsec:InstanceMatch}. 

\textbf{Dual-guided Modules.}
We introduce the Dual-guided Modules. This module enhances LiDAR BEV features and Camera BEV features at instance-level based on interpolation and neighborhood feature weighting, respectively, to facilitate the fusion of multi-modal BEV feature maps, as described in Sec.\ref{subsec:dg}. The Dual-guided Modules also represent our core innovation.

\textbf{Detection Head.}
Following \cite{bai2022transfusion}, we obtain the final 3D detection results. Details of the Detection Head and model constraints are provided in Sec.\ref{subsec:head}.

\subsection{Instance Match Modules}
\label{subsec:InstanceMatch}

\subsubsection{Instance-level Feature Generator (IFG)}
\label{subsubsec:ifg}

The Instance Match Modules first predict 3D proposals using BEV features, followed by score filtering and sampling to construct instance features. Prior to this, inspired by\cite{cao2019gcnet}, we refine $F_{C}$ to capture global long-range dependencies, as illustrated in Fig.\ref{fig:framework}.(b). For $F_{C}$, we apply a convolution layer with a $1 \times 1$ kernel to generate $F_{C}^{v}$ and $F_{C}^{k}$. After passing through a Softmax layer, $F_{C}^{k}$ is multiplied with $F_{C}^{v}$ via matrix multiplication, followed by element-wise addition with $F_{C}$ to obtain $F_{C}^{'}$. $F_{C}^{'}$ facilitates the subsequent construction of HIP, as it provides a richer global scene understanding compared to $F_{C}$.

Next, we extract instance-level features from both LiDAR and Camera modalities using a unified approach. For $F_{C}^{'}$ and $F_{L}$, we employ an additional prediction head \cite{Centerpoint} to obtain 3D proposals $P_{C} \in \mathbb{R}^{9}$ and $P_{L} \in \mathbb{R}^{9}$, along with their scores $S \in \mathbb{R}$. Proposals with scores below a threshold $\gamma$ are discarded. For the remaining proposals, we obtain the key sample set $[ \delta_{c}, \delta_{t}, \delta_{b}, \delta_{l}, \delta_{r}]$. Here, $\delta_{c}$ represents the center point coordinates of the proposal's 2D projection on the plane, while $\delta_{t}, \delta_{b}, \delta_{l}, \delta_{r}$ denote the midpoints of the four boundary lines of the 2D projection. The 2D coordinates of all key samples are directly or indirectly obtained from the predicted 3D proposal attributes $[x, y, z, w, h, l]$, where $x, y, z$ represent position, and $w, h, l$ denote the proposal's dimensions. Finally, we concatenate all key sample features to obtain instance-level features, as shown in Fig.\ref{fig:CameraBEV_Enhancement_and_Instance_Generation}. Note that the additional prediction head does not use the GT annotations during training or inference. Then, generate the camera instance $I_{C} \in \mathbb{R}^{N_{C} \times 1 \times 1 \times C_{C}}$ and the LiDAR instance $I_{L} \in \mathbb{R}^{N_{L} \times 1 \times 1 \times C_{L}}$, where $N_{C}$ and $N_{L}$ represent the number of camera instances and LiDAR instances.

\subsubsection{Difficulty-aware Instance Pair Matcher (DIPM)}
\label{subsubsec:dipm}

\begin{figure*}[t]
\centering
\includegraphics[width=1\textwidth]{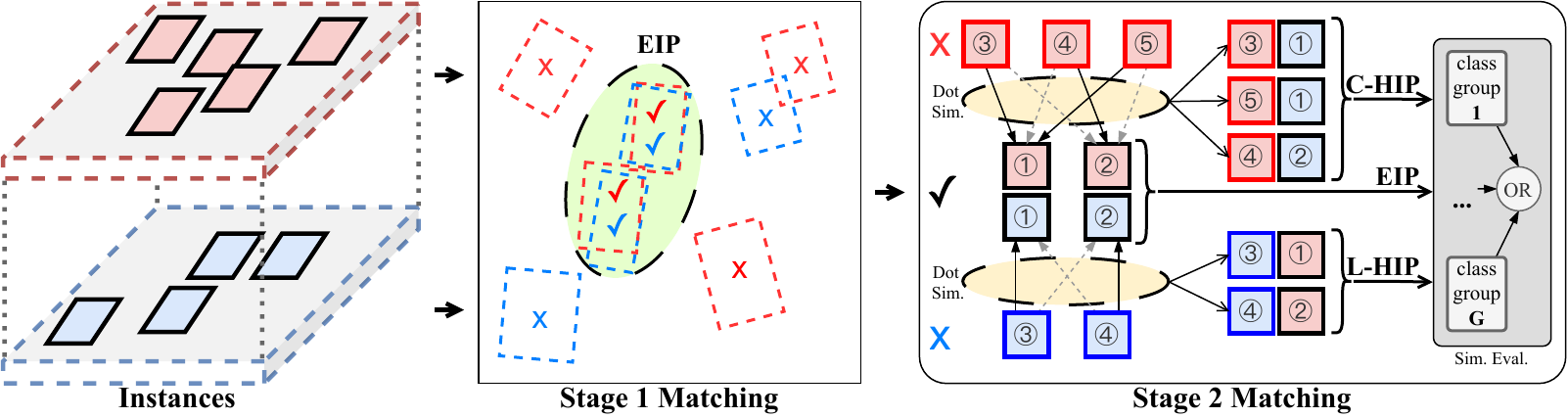}
\caption[ ]{Pipeline of \textbf{Difficulty-aware Instance Pair Matcher}. The DIPM operates in two sequential stages. Stage 1 selects cross-modally consistent easy instance pairs (EIPs) through LiDAR-dominated IoU matching, while Stage 2  constructs both camera-hard instance pairs (C-HIPs) and LiDAR-hard instance pairs (L-HIPs) by computing intra-modality feature similarities.} 
\label{fig:Instance_Match_Module_2}
\end{figure*}

High-performance perception components are required in real-world applications such as autonomous driving to improve safety performance. However, detecting distant, occluded, or small targets remains challenging. Unfortunately, few studies have explicitly focused on hard instance detection. As established in Sec.\ref{Sec:Introduction}, it is clear that `easy instances' are the primary focus of end-to-end detectors, although some hard instances exhibit distinct representations in multi-modal detectors. Therefore, after constructing the instance-level features, we propose the DIPM. DIPM simulates the process of identifying easy instances during inference while further distinguishing between two types of hard instances. The workflow of DIPM is shown in Fig.\ref{fig:Instance_Match_Module_2}.

The DIPM operates in a stage-wise manner. The target of Stage 1 (S1) is to identify instances that can be accurately detected across all modalities, which are labeled as easy instance pairs (EIP). For any LiDAR instance $I_L^i \in I_L$, $\forall i \in \{1, \dots, N_L\}$ and any camera instance $I_C^j \in I_C$, $\forall j \in \{1, \dots, N_C\}$ derived from a set of BEV features, we compute their Intersection over Union (IoU) in the 2D plane. The unique camera instance index $j_{i}$ for $I_L^i$ is determined by maximizing the IoU, formulated as:
\begin{equation}\label{eq_match_s1}
I_{C}^{j_{i}} = \arg \max_{} \text{ IoU}(I_{L}^{i}, I_{C}^{j})\mid j \in \{1, \dots, N_C\}
\end{equation}
where $I_{C}^{j_{i}}$ denotes the best-matched camera instance for $I_{L}^{i}$, and $IoU(.,.)$ represents the function for computing 2D IoU. To increase the confidence in the match, we use $\eta$ as the threshold for the IoU. The matching of S1 is LiDAR-instance-dominant, as easy instances are guaranteed to have a good representation across any modality. The LiDAR instance in the EIP provides sufficient information for the detection head. Finally, S1 outputs the EIP sequence $[\{I_{L}^{1},I_{C}^{1}\},...,\{I_{L}^{N_{E}},I_{C}^{N_{E}}\}], N^{E} \in \{0,min(N^{L},N^{C})\}$. 

In S1, all unmatched instances are considered potential targets, referred to as hard instances. The target of Stage~2 (S2) is to identify two types of Hard Instance Pairs (HIP): one is the C-HIP, which conforms to the `hard for Camera, easy for LiDAR' criterion, and the other is the L-HIP, which conforms to the `hard for LiDAR, easy for Camera' criterion. The construction processes for these two types of HIPs are symmetric. We take the construction of C-HIP as an example to illustrate the process. In S1, all camera instances are divided into two subsets: the successfully matched instance set $I_{C}^{m}$ (corresponding to the matched LiDAR instance set $I_{L}^{m}$), and the unmatched instance set $I_{C}^{um}$. For each instance $I_{C}^{um_{k}},\quad \forall k \in \{1, \dots, N_{C}-N_{E}\}$, we compute its dot-product similarity with all instances in $I_{C}^{m}$. The index $t$ of the easy camera instance for $I_{C}^{um_{k}}$ is then determined by finding the maximum dot product, which can be formulated as:
\begin{equation}\label{eq_I_C^{um}}
I_{C}^{m_{t}} = \arg \min_{} \text{ d}(I_{C}^{m}, I_{C}^{um_{k}})
\end{equation}
where $I_{C}^{m_{t}}$ represents the best-matched result for $I_{C}^{um_{k}}$, and $d(\cdot)$ denotes the dot product computation function. The corresponding easy LiDAR instance of $I_{C}^{m_{t}}$ is $I_{L}^{m_{t}}$. Consequently, the pair $I_{C}^{um_{k}}$ and $I_{L}^{m_{t}}$ is labeled as C-HIP. The matching process in S2 computes similarity between instances of the same modality, which aims to avoid errors introduced by cross-modal feature matching and improve confidence in the cross-modal matching results. We employ cosine similarity rather than dot product similarity as the constraint in S1 (see Sec.\ref{subsec:head}), with the objective of enhancing modality-invariant representation. Finally, S2 outputs two sequences: the C-HIP sequence $\left[\{I_{C}^{um_{1}},I_{L}^{m_{1}}\},...,\{I_{C}^{um_{N_{C}-N_{E}}},I_{L}^{m_{N_{C}-N_{E}}}\}\right]$ and the L-HIP sequence $\left[\{I_{L}^{um_{1}},I_{C}^{m_{1}}\},...,\{I_{L}^{um_{N_{L}-N_{E}}},I_{C}^{m_{N_{L}-N_{E}}}\}\right]$.

\begin{algorithm}[t!]
\renewcommand{\algorithmicrequire}{\textbf{Input:}}
\renewcommand{\algorithmicensure}{\textbf{Output:}}
\caption{\textbf{PGIE} (top for EIP and bottom for C-HIP)}
\label{alg:pgi}
\begin{algorithmic}[1]
\REQUIRE \;

$F_{C} \in \mathbb{R} ^{H\times W\times C_{C}}$: Camera BEV features\\

$[\{I_{L}^{1},I_{C}^{1}\},...,\{I_{L}^{N_{E}},I_{C}^{N_{E}}\}]$: EIP seq.\\ 

$[\{I_{C}^{um_{1}},I_{L}^{m_{1}}\},...,\{I_{C}^{um_{N_{C}-N_{E}}},I_{L}^{m_{N_{C}-N_{E}}}\}]$: C-HIP seq.\\

\STATE $F_{C}^{Enh}$ $\leftarrow$ \textbf{Clone}($F_{C}$)

\STATE EIP $\leftarrow$ \textbf{Point\_Squeeze}(EIP$, C_{C}$)

\STATE $F_{C}^{temp1}$ $\leftarrow$ \textbf{Expand}(\textbf{Clone}($F_{C}$)$, N_{E}, H, W, C_{C}$)

\FOR {$idx \in (1, N_{E})$}

    \STATE  $coord$ $\leftarrow$ \textbf{Absl\_to\_Rela}$(x_{ctr}(I_{C}^{idx}), y_{ctr}(I_{C}^{idx}))$
    
    \STATE ${gs}^{idx}$ $\leftarrow$ \textbf{Bilinear}$(F_{C}^{temp1}[idx],coord)$

    \STATE $gs_{Enh}^{idx} \leftarrow {gs}^{idx}$ $\bigotimes I_{L}^{idx}$
    
    \STATE $F_{C}^{Enh} \leftarrow F_{C}[coord[0],coord[1],:]$  $\bigoplus$ $gs_{Enh}^{idx}$

\ENDFOR

\dashedline

\STATE C-HIP $\leftarrow$ \textbf{Point\_Squeeze}(C-HIP$, C_{C}$)

\STATE $F_{C}^{temp2}$ $\leftarrow$ \textbf{Expand}(\textbf{Clone}($F_{C}$)$, N_{C}-N_{E}, H, W, C_{C}$)

\FOR {$idx \in (1, N_{C}-N_{E})$}

    \STATE  $coord$ $\leftarrow$ \textbf{Absl\_to\_Rela}($x_{ctr}(I_{C}^{um_{idx}}), y_{ctr}(I_{C}^{um_{idx}})$)
    
    \STATE ${gs}^{idx}$ $\leftarrow$ \textbf{Bilinear}($F_{C}^{temp2}[idx],coord$)

    \STATE $gs_{Enh}^{idx} \leftarrow  {gs}^{idx}$ $\bigotimes I_{L}^{m_{idx}}$
    
    \STATE $F_{C}^{Enh} \leftarrow F_{C}^{Enh}[coord[0],coord[1],:]$  $\bigoplus$ $gs_{Enh}^{idx}$

\ENDFOR

\ENSURE \;
    
$F_{C}^{Enh}$: Enhanced Camera BEV feature.
\end{algorithmic}
\end{algorithm}

\begin{algorithm}[t!]
\renewcommand{\algorithmicrequire}{\textbf{Input:}}
\renewcommand{\algorithmicensure}{\textbf{Output:}}
\caption{\textbf{IGPE}}
\label{alg:igp}
\begin{algorithmic}[1]
\REQUIRE \;

$F_{L} \in \mathbb{R} ^{H\times W\times C_{L}}$: LiDAR BEV features\\

$\{I_{L}^{um_{1}},I_{C}^{m_{1}}\},...,\{I_{L}^{um_{N_{L}-N_{E}}},I_{C}^{m_{N_{L}-N_{E}}}\}]$: L-HIP seq.\\

\STATE $F_{L}^{Enh}$ $\leftarrow$ \textbf{Clone}($F_{L}$)

\STATE L-HIP $\leftarrow$ \textbf{Point\_Excitation}(L-HIP$, C_{L}$)

\STATE $dist$ $\leftarrow$ [\textbf{D}$(I_{L}^{um_{idx}},I_{C}^{m_{idx}})$ \textbf{for} $idx \in N_{L}-N_{E}$]

\STATE $w \leftarrow$ [$1 - \frac{d - \textbf{Min}(dist)}{\textbf{Max}(dist) - \textbf{Min}(dist)}$ \textbf{for} $d \in dist$]

\FOR {$idx \in (1, N_{L}-N_{E})$}

    \STATE  $coord$ $\leftarrow$ \textbf{Absl\_to\_Rela}($x_{ctr}(I_{L}^{um_{idx}}), y_{ctr}(I_{L}^{um_{idx}})$)

    \STATE $[\delta_\nwarrow, \delta_\swarrow, \delta_\nearrow, \delta_\searrow]$ $\leftarrow$ \textbf{Surr\_Int\_Coord}($coord$)

    \STATE $I_{C}^{{mw}_{idx}} \leftarrow$ $I_{C}^{m_{idx}} \bigotimes w_{idx}$

    \FOR {$\delta \in [\delta_\nwarrow, \delta_\swarrow, \delta_\nearrow, \delta_\searrow]$}

        \STATE $F_{L}^{Enh} \leftarrow F_{L}[{\delta}[0],{\delta}[1],:] \bigoplus$ $I_{C}^{{mw}_{idx}}$

    \ENDFOR
    
\ENDFOR

\ENSURE \;
    
$F_{L}^{Enh}$: Enhanced LiDAR BEV feature.
\end{algorithmic}
\end{algorithm}

We further evaluate whether all EIP and HIP pairs are reasonable matches. One key criterion we consider is the similarity in classification categories \cite{CBGS, peri2023towardsLT3D}. Namely, if two instance-level features in an EIP or HIP exhibit high similarity in their categories, we consider them a valid match; otherwise, the pair is invalidated. We propose two category similarity grouping strategies for different data scales and conduct thorough experimental analysis to demonstrate that the instance matching strategy and the evaluation strategy for instance pairs have a significant positive impact on the fusion process.

\subsection{Dual-guided Modules}
\label{subsec:dg}
The Dual-guided Modules aim to achieve two types of guided enhancement for BEV features using EIP and HIP. The first one employs EIPs and C-HIPs to perform point-guided-image enhancement for the camera BEV features, and the second utilizes L-HIP to conduct image-guided-point enhancement for the LiDAR BEV features.

\subsubsection{Point-guide-Image Enhancement}
\label{subsubsec:pgie}
The PGIE utilizes the easy LiDAR instances from either EIP or C-HIP to guide their corresponding camera instances. The overall process is outlined in Algorithm.\ref{alg:pgi}. Since the guided enhancement procedure for EIP is identical to that for C-HIP, we use the former as an illustrative example. 

First, we employ a linear layer to unify the feature dimensions of cross-modal features in EIP to $C_{C}$. Subsequently, we expand the dimensions of $F_{C}$ and denote it as $F_{C}^{temp1}$, which is then used for interpolation operations. For any instance feature pair $\{I_{L}^{idx}, I_{C}^{idx}\}$ from the EIP sequence, we transform the 2D center coordinates of the camera instance from absolute coordinates (representing positions in the actual detection range) to relative coordinates (representing positions on BEV features). 

Next, a grid sample operation relocates the original features at these coordinates to ${gs}_{idx}$ via bilinear interpolation. The enhanced local features are then obtained through element-wise multiplication between the LiDAR instance and ${gs}_{idx}$, followed by element-wise addition at the corresponding positions of the camera instance in $F_{C}$. We then repeat this operation using C-HIPs. The final output is the enhanced Camera BEV feature $F_{C}^{Enh}$. PGIE inherits and amplifies the inherent advantages of LiDAR-dominant approaches.

\subsubsection{Image-guide-Point Enhancement}
\label{subsubsec:igpe}
The IGPE employs the easy camera instances from L-HIP to guide their corresponding hard LiDAR instances. The complete procedure is summarized in Algorithm.\ref{alg:igp}. 

For each instance pair in the HIP, we first compute their Euclidean distance and utilize the normalized distance as the feature fusion weight $w$. This weight $w$ can be interpreted as IGPE's preference for different levels of instance pairs. Within the same batch of data, instance pairs with smaller distances are considered to have higher confidence.

Next, for any instance feature pair $\{I_{L}^{um_{idx}}, I_{C}^{m_{idx}}\}$ from the L-HIP sequence, we perform coordinate transformation on the center coordinates of $I_{L}^{um_{idx}}$. Subsequently, we obtain four integer coordinates surrounding the non-integer center coordinates: $[\delta_\nwarrow, \delta_\swarrow, \delta_\nearrow, \delta_\searrow]$.

Finally, the result of element-wise multiplication between $I_{C}^{m_{idx}}$ and $w$ is added to $F_{L}$ at these four integer coordinates through element-wise addition, generating the enhanced LiDAR BEV feature $F_{L}^{Enh}$. The proposed IGPE significantly improves the discriminative power of hard LiDAR instances, resulting in BEV features with more robust semantic information after fusion.

\subsection{Detection Head and Loss Function}
\label{subsec:head}

We concatenate $F_{L}^{E}$ and $F_{C}^{E}$ along the channel dimension to generate the fused BEV feature, and then, following TransFusion\cite{bai2022transfusion}, produce the final 3D detection results. Focal loss \cite{lin2017focal} and L1 loss are applied to classification and 3D bounding box regression, respectively. Additionally, we observe the occurrence of an empty list for EIP during the early stages of training, which is caused by both our instance filtering strategy and the low-quality parameters of the additional detection head in the early training phase. Therefore, we apply focal loss to the additional detection head and use cosine similarity loss during the PGIE to further constrain the network. The cosine similarity loss $\mathcal{L}_{Cos}$ is defined as:
\begin{equation}
\mathcal{L}_{Cos} = \frac{1}{N_E} \sum_{i=1}^{N_E} \left( 1 - \frac{I_L^{i} \cdot I_C^{i}}{\| I_L^{i} \| \| I_C^{i} \|} \right)
\end{equation}
where $N_{E}$ is defined in Sec.\ref{subsec:InstanceMatch}, and the $i$-th pair of instance-level features in the EIP sequence is represented as $\{I_{L}^{i}, I_{C}^{i}\}$. The network's loss $\mathcal{L}$ is defined as:
\begin{equation}
\mathcal{L} =\lambda_1 \mathcal{L}_{H}+\lambda _2 \mathcal{L}_{L}+\lambda_3 \mathcal{L}_{C}+\lambda_4 \mathcal{L}_{Cos}
\end{equation}
where $\mathcal{L}_{H}$, $\mathcal{L}_{L}$, $\mathcal{L}_{C}$, and $\mathcal{L}_{Cos}$ represent the loss of the final Detection Head, the instance prediction loss of the LiDAR branch, the instance prediction loss of the camera branch, and the cosine similarity loss, respectively. ${\lambda}_{1}$, ${\lambda}_{2}$, ${\lambda}_{3}$, and ${\lambda}_{4}$ are the weight coefficients for the respective losses. During training, PGIE dynamically maintains the historical maximum of the cosine similarity loss (initialized to 0) and applies it to each batch where the EIP is empty.

\section{Experiments}
\label{Sec:Experiments}

\subsection{Details}
\label{subsec:Details}

\textbf{Dataset and Evaluation Metrics.}
The nuScenes dataset\cite{nuscenes} is a large-scale outdoor dataset comprising 1,000 multi-modal scenes. The LiDAR data and Camera data we use are collected from a top-mounted 32-beam LiDAR sensor and six-view Camera sensors, respectively. Our primary evaluation metrics are derived from the official nuScenes metrics, including the mean Average Precision (mAP) across 10 common categories and the nuScenes Detection Score (NDS). mAP is one of the most commonly used evaluation metrics in object detection algorithms\cite{mao20233d_zongshu,zou2023object_2dsurvey}, measuring the average precision of the model across multiple categories, considering both the accuracy and recall of the detection results. NDS is a comprehensive metric specific to the nuScenes dataset, combining mAP with other key factors such as velocity, size, and orientation into a weighted score, providing a more holistic evaluation of the detection performance in autonomous driving models.

\textbf{Implementation.}
We implement DGFusion on the open-source platform OpenPCDet\cite{openpcdet} and PyTorch\cite{paszke2019pytorch}. In the LiDAR branch, the voxel size is set to $[0.075m, 0.075m, 0.2m]$, and the point cloud range is defined along the $X$, $Y$, and $Z$ axes as $[-54m, -54m, -5m, 54m, 54m, 3m]$, with TransFusion-L\cite{bai2022transfusion} used as the pre-trained model. For the Camera branch, under the LSS\cite{lift_splat_shoot} configuration, the frustum ranges are set to $[-54m, 54m, 0.3m]$ for the $X$ axis, $[-54m, 54m, 0.3m]$ for the $Y$ axis, and $[-10m, 10m, 20m]$ for the $Z$ axis, with the depth range set to $[1m, 60m, 0.5m]$. Swin Transformer\cite{Swintransformer} is employed as the pre-trained model. For both branches, the prediction head for generating additional proposals follows the same configuration as CenterPoint\cite{Centerpoint}.

\textbf{Training and Inference.}
Our baseline models (in     Table.\ref{tab_nuscenes_test}, Table.\ref{tab_nuscenes_val}, Table.\ref{tab_robustness_range+vis+size},  Table.\ref{tab_ablation_latency} and Fig.\ref{fig:recall}) and the models trained on small-scale datasets (in Table.\ref{tab_robustness_samll_scale}, Table.\ref{tab_ablation_component}, Table.\ref{tab_ablation_instance_pairs_evaluation}, Table.\ref{tab_ablation_instance_feature_type} and Table.\ref{tab_ablation_gamma}) trained using 8 NVIDIA GeForce RTX 3090 24G GPUs, with a total batch size of 16. During training, data augmentation techniques are applied, including random flipping, rotation (within the range[$-\frac{\pi}{4}$, $\frac{\pi}{4}$]), translation (standard deviation=0.5), LiDAR data scaling, and camera data scaling. Additionally, CBGS\cite{CBGS} is employed for re-sampling the training data. The training process utilizes the Adam optimizer with a cosine annealing learning rate schedule, with an initial learning rate of 0.0001 and a weight decay of 0.01. $\gamma$ is set to $0.7$, $\eta$ is set to $0.7$, and $\lambda_{1},\lambda_{2},\lambda_{3},\lambda_{4}$ are set to $0.99, {10}^{-4}, {10}^{-4}, {10}^{-2}$. No test-time augmentation (TTA) is applied. 

\subsection{Main Results}

\subsubsection{Results on NuScenes Test Set}
As shown in Table \ref{tab_nuscenes_test}, we compare DGFusion with other models on the nuScenes test benchmark. BEVFusion\cite{bevfusion-mit} serves as our baseline. Apart from BEVFusion, these approaches cover single-modal methods\cite{ji2024enhancing,jiang2024far3d,CBGS,Centerpoint,bai2022transfusion,liu2024sparsedet,chen2023focalformer3d,hou2025polarbevu,wang2025rethinking}, multi-modal methods\cite{mvp,peri2023towardsLT3D,ma2023long_LT3D_2dlatefusion,graphalign,graphalign++,bai2022transfusion,bevfusion-pku,UniTR,ObjectFusion,chen2023focalformer3d,cao2024tfienet}, solutions addressing hard instances or robustness\cite{jiang2024far3d,liu2024sparsedet,peri2023towardsLT3D,ma2023long_LT3D_2dlatefusion,graphalign++,UniTR,chen2023focalformer3d,cao2024tfienet}, the Point-guide-Image paradigm\cite{graphalign,graphalign++,bai2022transfusion,bevfusion-pku,UniTR,ObjectFusion,chen2023focalformer3d,cao2024tfienet}, the Image-guide-Point paradigm\cite{mvp,peri2023towardsLT3D,ma2023long_LT3D_2dlatefusion,qi2018frustum_pointnets,wang2020high_7Dpointnet,pointpainting,wang2021pointaugmenting}, and approaches using similar techniques to our DGFusion\cite{CBGS,bai2022transfusion,ObjectFusion}. 

Firstly, DGFusion achieves 71.2\% mAP and 73.7\% NDS on the nuScenes test benchmark, improving 1.0\% mAP and 0.8\% NDS over the baseline BEVFusion\cite{bevfusion-mit}. DGFusion shows notable improvements of 1.2\%, 1.3\%, 1.4\%, and 2.1\% in Bus, Motorcycle, Bike, and Traffic Cone categories, respectively. DGFusion outperforms all single-modality methods, indicating that multi-modal fusion is an ideal solution for mitigating hard instance detection issues. 

Secondly, compared to the multi-modal version of FocalFormer3D\cite{chen2023focalformer3d}, DGFusion shows improvements in Car, Construction Vehicle, Barrier, and Pedestrian categories. Although DGFusion lags slightly behind  FocalFormer3D\cite{chen2023focalformer3d} in terms of mAP and NDS, its inference latency is significantly lower (see Table \ref{tab_ablation_latency}), demonstrating DGFusion's advantages in practical scenarios. 

Lastly, compared to instance-based ObjectFusion\cite{ObjectFusion}, DGFusion improves mAP by 0.2\% and NDS by 0.4\%. Additionally, we observe that within multi-modal methods, the Image-guide-Point approach\cite{peri2023towardsLT3D,ma2023long_LT3D_2dlatefusion} shows significant advantages in Truck and Bike categories. This provides valuable insights for further exploring hard instance detection.

\subsubsection{Results on NuScenes Validation Set}

As shown in Table.\ref{tab_nuscenes_val}, DGFusion achieves 69.2\% mAP and 71.8\% NDS on the nuScenes validation benchmark, surpassing the baseline method BEVFusion \cite{bevfusion-mit} by 0.7\% mAP and 0.5\% NDS. DGFusion shows significant improvements in categories with a high prevalence of hard instances, such as Bike, Pedestrian, and Traffic Cones.

\begin{table*}[t]
\scriptsize
\centering
\caption{Comparison with the SOTA methods on the nuScenes \textbf{test} set.}
\renewcommand\arraystretch{0.9}
\tabcolsep=1.8mm 
\resizebox{\linewidth}{!}{
\begin{tabular}{l|cc|cc|cccccccccc}
\toprule
Method & Modality & Paradigm & mAP  & NDS  & Car & Truck & C.V. & Bus & Trailer & Barrier & Motor. & Bike & Ped. & T.C. \\ 
\midrule
\midrule
PolarBEVU\cite{hou2025polarbevu}&C& - & 49.9& 57.4& -& -& -& -& - &-& - &- &-& -\\
QAF2D\cite{ji2024enhancing}&C& - & 56.6& 64.2& -& -& -& -& - &-& - &- &-& -\\
Far3D\cite{jiang2024far3d}&C& - & 63.5& 68.7& -& -& -& -& - &-& - &- &-& -\\
CBGS\cite{CBGS}& L& - & 52.8& 63.3& 81.1& 48.5& 10.5& 54.9& 42.9& 65.7& 51.5& 22.3& 80.1& 70.9\\
CenterPoint\cite{Centerpoint} &L& - & 58.0& 65.5 & - & -& - & - & - & -& -& - & -& -\\
TransFusion-L \cite{bai2022transfusion}& L& - & 65.5& 70.2 &86.2& 56.7& 28.2& 66.3& 58.8& 78.2& 68.3& 44.2& 86.1& 82.0\\
SparseDet\cite{liu2024sparsedet}& L& - & 66.7& 71.9& 86.2& 56.0& 30.2& 66.5& 58.4& 78.7& 73.7& 46.8& 87.5& 82.5\\
LDFCNet\cite{wang2025rethinking}& L& - & 67.9 &72.2& 85.3& 54.7& 66.6& 62.6& 29.6& 88.6& 76.3& 52.0& 86.1& 77.5\\
FocalFormer3D-L \cite{chen2023focalformer3d} &L& - & 68.7& 72.6& 87.2& 57.1& 34.4& 69.6& 64.9 &77.8& 76.2 &49.6 & 88.2& 82.3\\
\midrule
\midrule

  $\ast$Frustum PointNet\cite{qi2018frustum_pointnets}& LC & IGP& 32.5 & 42.7 & 44.0 & 23.0 & 11.0 & 24.0  & 11.0 & 43.0 & 41.0 & 27.0 & 54.0 & 46.0  \\
  7D PointNet\cite{wang2020high_7Dpointnet}& RLC & IGP& 36.6 & 46.8 & 48.0 & 29.0 & 23.0 & 33.0  & 18.0 & 44.0 & 41.0 & 28.0 & 56.0 & 48.0  \\
  PointPainting\cite{pointpainting}& LC & IGP & 46.4 & 58.1 & 77.9 & 35.8  & 15.8 & 36.2 & 37.3 & 60.2 & 41.5 & 24.1 & 73.3 & 62.4 \\
  PointAugmenting\cite{wang2021pointaugmenting}& LC& IGP& 66.8& 71.0& 87.5& 57.3& 28.0& 65.2& 60.7& 72.6& 74.3& 50.9& 87.9& 83.6\\
  PAI3D\cite{liu2022pai3d}& LC&IGP& 71.4& 74.2& 88.4& 62.7& 37.8& 71.3& 65.8& 75.5& 80.8& 58.2& 90.3& 83.2\\
MVP\cite{mvp}& LC & IGP & 66.4 & 70.5 & 86.8 & 58.5  & 26.1 & 67.4 & 57.3 & 74.8 & 70.0   & 49.3 & 89.1 & 85.0 \\
LT3D\cite{peri2023towardsLT3D}& LC & IGP & -& -& 88.5& 63.4& 29.0& -& -& -& 68.2& 58.5& -& - \\
Y. Ma {\it et al.}\cite{ma2023long_LT3D_2dlatefusion}& LC & IGP & -& -& 86.3& 60.6& 35.3& -& -& -& 75.9& 70.1& -& - \\

\midrule
GraphAlign\cite{graphalign}& LC & PGI  &66.5&70.6& 87.6& 57.7& 26.1& 66.2& 57.8& 74.1& 72.5& 49.0& 87.2 &86.3\\
GraphAlign++\cite{graphalign++}& LC & PGI  &68.5&72.2& 87.5& 58.5& 32.3& 68.9& 58.3& 74.3& 76.4& 53.9& 88.3 &86.3\\
TFIENet\cite{cao2024tfienet}& LC & PGI  &-&-& 86.4& 61.8& -& 70.6& 60.5& 79.5& -& 55.6& 88.2 &-\\
TransFusion\cite{bai2022transfusion}& LC & PGI & 68.9 & 71.7 & 87.1 & 60.0 & 33.1 & 68.3 & 60.8 & 78.1 & 73.6 & 52.9 & 88.4 & 86.7\\
BEVFusion(PKU)\cite{bevfusion-pku}& LC & PGI & 69.2 & 71.8 & 88.1 & 60.9 & 34.4 & 69.3 & 62.1 & 78.2 & 72.2 & 52.2 & 89.2 & 85.2\\
UniTR\cite{UniTR} &LC& PGI & 70.9&74.5&87.9& 60.2& 39.2&72.2&65.1&76.8&75.8&52.2&89.4&89.7\\
ObjectFusion\cite{ObjectFusion}& LC & PGI &71.0& 73.3& 89.4& 59.0 & 40.5 & 71.8 & 63.1& 76.6 & 78.1 & 53.2 & 90.7 & 87.7\\
FocalFormer3D\cite{chen2023focalformer3d}&LC& PGI & 71.6& 73.9 &88.5&61.4&35.9&71.7&66.4&79.3&80.3&57.1& 89.7& 85.3\\
\midrule
\midrule
BEVFusion  \cite{bevfusion-mit}& LC & PGI & 70.2 & 72.9 & 88.6 & 60.1 & 39.3& 69.8 & 63.8 & 80.0 & 74.1 & 51.0 & 89.2 & 86.5\\ 
\textbf{DGFusion(Ours)}& LC & \textbf{DG} &71.2 &73.7 &88.8 &60.7 &40.1 &71.0 &64.5 &80.6 &75.4 &52.4 &90.1 &88.6 \\
\rowcolor{red!20} \textcolor{red}{+}& &
&\textit{\textcolor{red}{+1.0}}
&\textit{\textcolor{red}{+0.8}} 
&\textit{\textcolor{red}{+0.2}}
&\textit{\textcolor{red}{+0.6}}
&\textit{\textcolor{red}{+0.8}}
&\textit{\textcolor{red}{+1.2}}
&\textit{\textcolor{red}{+0.7}}
&\textit{\textcolor{red}{+0.6}}
&\textit{\textcolor{red}{+1.3}}
&\textit{\textcolor{red}{+1.4}}
&\textit{\textcolor{red}{+0.9}}
&\textit{\textcolor{red}{+2.1}}
\\
\bottomrule
\end{tabular} }
\begin{tablenotes}  
\footnotesize  
\item[1] [1] `C.V.', `Motor.', `Ped.', and `T.C.' are short for construction vehicle, motorcycle, pedestrian, and traffic cone, respectively.  
\item[2] [2] `L' means only LiDAR data are used, `C' means only Camera data are used, `LC' denotes the use of both LiDAR and Camera data, `RLC' denotes the use of Radar, LiDAR and Camera data. `IGP' means the Image-guided-Point paradigm, `PGI' means the Point-guided-Image paradigm, and `DG' means the Dual-guided paradigm.  
\item[1] [3] $\ast$ results are cited from \cite{wang2020high_7Dpointnet}.
\end{tablenotes}
\label{tab_nuscenes_test}
\end{table*}

\begin{table*}[t]
\scriptsize
\centering
\caption[ ]{Comparison with the SOTA methods on the nuScenes  \textbf{val.} set.}
\renewcommand\arraystretch{0.9}
\tabcolsep=1.8mm
\resizebox{\linewidth}{!}{
\begin{tabular}{l|cc|cc|cccccccccc }
\toprule
Method & Modality & Paradigm & mAP & NDS & Car & Truck & C.V. & Bus  & Trailer & Barrier & Motor. & Bike & Ped. & T.C. \\ 
\midrule
\midrule
PolarBEVU\cite{hou2025polarbevu}&C& - & 49.1& 56.5& -& -& -& -& - &-& - &- &-& -\\
QAF2D\cite{ji2024enhancing}&C& - & 50.0& 58.6& -& -& -& -& - &-& - &- &-& -\\
Far3D\cite{jiang2024far3d}&C& - & 51.0& 59.4& -& -& -& -& - &-& - &- &-& -\\
TransFusion-L \cite{bai2022transfusion} &L & - & 65.1 &70.1& 86.5& 59.6& 25.4& 74.4& 42.2& 74.1& 72.1& 56.0& 86.6& 74.1\\
SparseDet\cite{liu2024sparsedet}& L& - & 65.3& 70.3& 87.5& 60.2& 27.2& 75.8& 40.4& 73.1& 69.7& 58.8& 86.5& 63.4\\
LDFCNet\cite{wang2025rethinking}& L& - & 66.2& 71.0& 86.4& 60.0& 73.5& 43.6& 24.0& 88.9& 75.7& 61.9& 78.9& 68.6\\
FocalFormer3D-L \cite{chen2023focalformer3d} &L & - & 66.5 &71.1& -& -&-&-&-&-&-&-&-&-\\
\midrule
\midrule
  7D PointNet\cite{wang2020high_7Dpointnet}& RLC & IGP& 46.8 & 52.3 & 45.0 & 39.0 & 41.0 & 34.0  & 13.0 & 49.0 & 51.0 & 46.0 & 70.0 & 70.0  \\
  PAI3D\cite{liu2022pai3d}& LC&IGP& 67.6& 71.1& -& -&-&-&-&-&-&-&-&-\\
\midrule
GraphAlign++\cite{graphalign++}& LC & PGI  &69.9&73.1& 89.1& 63.2& 33.1& 75.1& 67.8& 74.1& 73.0& 49.9& 88.2 &86.7\\
TransFusion \cite{bai2022transfusion}&LC& PGI &  67.3 &71.2& 87.6& 62.0& 27.4& 75.7& 42.8& 73.9& 75.4& 63.1& 87.8& 77.0\\ 
BEVFusion-PKU  \cite{bevfusion-pku} &LC & PGI & 67.9& 71.0& 88.6& 65.0& 28.1& 75.4& 41.4& 72.2& 76.7& 65.8& 88.7& 76.9\\
UniTR\cite{UniTR} &LC& PGI & 70.0&73.1&-& -& -&-&-&-&-&-&-&-\\
ObjectFusion  \cite{ObjectFusion} &LC& PGI &69.8& 72.3 &89.7& 65.6& 32.0& 77.7& 42.8& 75.2& 79.4& 65.0&89.3& 81.1\\
\midrule
\midrule
BEVFusion  \cite{bevfusion-mit}&LC & PGI & 68.5 & 71.4  & 89.2 & 64.6 & 30.4 & 75.4 & 42.5 & 72.0 & 78.5 & 65.3 & 88.2 & 79.5\\
\textbf{DGFusion(Ours)}& LC & \textbf{DG} &69.2 &71.9 &89.3 &65.5 &30.7 &76.4 &43.2 &72.1 &79.0 &66.3 &89.0 &80.8 \\
\rowcolor{red!20} \textcolor{red}{+}& & 
&\textit{\textcolor{red}{+0.7}}
&\textit{\textcolor{red}{+0.5}} 
&\textit{\textcolor{red}{+0.1}}
&\textit{\textcolor{red}{+0.9}}
&\textit{\textcolor{red}{+0.3}}
&\textit{\textcolor{red}{+1.0}}
&\textit{\textcolor{red}{+0.7}}
&\textit{\textcolor{red}{+0.1}}
&\textit{\textcolor{red}{+0.5}}
&\textit{\textcolor{red}{+1.0}}
&\textit{\textcolor{red}{+0.8}}
&\textit{\textcolor{red}{+1.3}}
\\
\bottomrule
\end{tabular} }
\begin{tablenotes}  
\footnotesize  
\item[1] [1] `C.V.', `Motor.', `Ped.', and `T.C.' are short for construction vehicle, motorcycle, pedestrian, and traffic cone, respectively.
\item[2] [2] `L' means only LiDAR data are used, `C' means only Camera data are used, `LC' denotes the use of both LiDAR and Camera data, `RLC' denotes the use of Radar, LiDAR and Camera data. `IGP' means the Image-guided-Point paradigm, `PGI' means the Point-guided-Image paradigm, and `DG' means the Dual-guided paradigm. 
\end{tablenotes}
\label{tab_nuscenes_val}
\end{table*}

\subsection{Robustness Analysis}

\subsubsection{Hard Instances Evaluation}

\begin{table*}[t]
\scriptsize
\centering
\caption{Evaluation across ego-vehicle distance, visibility level, and object size on the nuScenes val. dataset.}
\renewcommand\arraystretch{0.9}
\tabcolsep=1.5mm
\resizebox{\linewidth}{!}{
\begin{tabular}{l|cc|cc|cc|cc|cc|cc|cc|cc}
\toprule
\multirow{4}{*}{Method}& \multicolumn{6}{c|}{Distance-Specific Evaluation}& \multicolumn{4}{c|}{Visibility-Specific Evaluation} &\multicolumn{6}{c}{Size-Specific Evaluation}\\ 
\cmidrule(lr){2-17} 
&\multicolumn{2}{c|}{0-20 $m$}  & \multicolumn{2}{c|}{20-40 $m$} & \multicolumn{2}{c|}{40+ $m$} & \multicolumn{2}{c|}{token=4} & \multicolumn{2}{c|}{token=3/2/1} &\multicolumn{2}{c|}{0-10$ m^3$} &\multicolumn{2}{c|}{10-30 $m^3$} &\multicolumn{2}{c}{30+ $m^3$} \\
\cmidrule(lr){2-17}
&mAP&NDS&mAP&NDS&mAP&NDS&mAP&NDS&mAP&NDS&mAP&NDS&mAP&NDS&mAP&NDS\\
\midrule
\midrule
BEVFusion\cite{bevfusion-mit}&79.5 &78.4 &60.1 &66.2 &33.4 &48.9 &53.2&64.0&14.0&42.2&37.6&43.0&17.3&24.3&23.2&31.7\\
\textbf{DGFusion(Ours)}&79.8 &78.5 &61.0 &66.6 &34.6 &49.5 &53.9&64.5&14.8&42.5&38.1&43.4&17.5&24.4&23.4&31.9\\
\rowcolor{red!20} \textit{\textcolor{red}{+}}
&\textit{\textcolor{red}{+0.3}}&\textit{\textcolor{red}{+0.1}}
&\textit{\textcolor{red}{+0.9}}&\textit{\textcolor{red}{+0.4}} 
&\textit{\textcolor{red}{+1.2}}&\textit{\textcolor{red}{+0.6}} 
&\textit{\textcolor{red}{+0.7}}&\textit{\textcolor{red}{+0.5}}
&\textit{\textcolor{red}{+0.8}}&\textit{\textcolor{red}{+0.3}} 
&\textit{\textcolor{red}{+0.5}}&\textit{\textcolor{red}{+0.4}} 
&\textit{\textcolor{red}{+0.2}}&\textit{\textcolor{red}{+0.1}}
&\textit{\textcolor{red}{+0.2}}&\textit{\textcolor{red}{+0.2}}\\
\bottomrule
\end{tabular}
}
\label{tab_robustness_range+vis+size}
\end{table*}

To demonstrate that our DGFusion mitigates the challenges associated with hard instance detection, we conduct segmented experiments from three perspectives: ego-vehicle distance, visibility level, and object volume. The results are presented in Table.\ref{tab_robustness_range+vis+size}. All evaluations were implemented based on the nuScenes DevKit\cite{nuscenes}. Specifically, the Visibility-Specific Evaluation involves masking GT annotations with different Visibility Tokens during the inference process. Similarly, the Range-Specific Evaluation and Size-Specific Evaluation involve masking prediction annotations.

First, we segment the predicted bounding boxes into three distance ranges relative to the ego-vehicle: 0-20$m$, 20-40$m$, and 40+$m$, and evaluated them separately. The basis for this range division strategy is twofold: 1) given the maximum range of point clouds (54m, see Sec.\ref{subsec:Details}), the division method with an interval of 20m is approximately equivalent to dividing all objects into three equal parts by distance, and 2) it refers to the experimental designs in \cite{liu2024sparsedet,song2024graphbev}. Therefore, these three ranges can be regarded as Near, Middle, and Far in sequence. DGFusion outperforms BEVFusion across all three distance ranges. Specifically, DGFusion achieves mAPs of 79.8\%, 61.0\%, and 34.6\% for the 0-20$m$, 20-40$m$, and 40+$m$ ranges, respectively, representing improvements of 0.3\%, 0.9\%, and 1.2\% over BEVFusion. Notably, DGFusion exhibits significant enhancements in the 20-40$m$ and 40+$m$ ranges, demonstrating its superior robustness in detecting distant objects.

Second, we evaluate the baseline model and our DGFusion based on the Visibility Token. The Visibility Token, an integral component of nuScenes GT annotations, quantifies the visibility level of instances within the camera sensor data. It comprises four levels: level `1' indicates visibility ranging from 0\% to 40\%, level `2' from 40\% to 60\%, level `3' from 60\% to 80\%, and level `4' from 80\% to 100\%. DGFusion achieves improvements in mAP for both high-visibility (token=4) and low-visibility (token=3/2/1) targets, with gains of 0.7\% and 0.8\%, respectively. The involvement of camera instances in both the EIP and HIP construction processes of DGFusion enhances the model's tolerance to instances with significant information density gaps.
 
Finally, we analyze the impact of object size on the baseline model and DGFusion. We define object size as the product of the length, width, and height of the GT bounding box (in $m^3$). We calculated the size ranges of all categories in the nuScenes training set, with representative results as follows: 1) Traffic Cone = (0.01, 1.69), 2) Truck = (8.95, 361.41), 3) Car = (1.80, 158.30), 4) Pedestrian = (0.07, 5.63), 5) Barrier = (0.05, 12.08), 6) Motorcycle = (0.56, 16.01), 7) Bicycle = (0.21, 5.73), 8) Bus = (21.83, 410.22), 9) Trailer = (3.16, 1504.15), and 10) C.V. = (1.06, 751.26). It can be seen that 0-10$m^3$, which includes Pedestrian, Bicycle, and Traffic Cones, is regarded as Small. 10-30$m^3$ applies to all categories except Pedestrian and Bicycle, and is regarded as Medium. 30+$m^3$ is more suitable for categories with larger size upper limits, such as Truck, Bus, Trailer, and C.V., and is regarded as Large. Compared to BEVFusion, our DGFusion consistently enhances performance across all object sizes, further narrowing the performance gap. Moreover, DGFusion demonstrates significant performance improvements for small objects, with a 0.5 increase in mAP and a 0.4 increase in NDS. In conclusion, our DGFusion exhibits greater robustness to variations in ego-vehicle distance, visibility level, and object size.

\subsubsection{Small-Scale Evaluation}

\begin{table*}[t]
\scriptsize
\centering
\caption[ ]{Training on Small-Scale sets and evaluation on the nuScenes val. set.}
\renewcommand\arraystretch{0.9}
\tabcolsep=0.6mm
\resizebox{\linewidth}{!}{
\begin{tabular}{l|l|cc|cccc|cccccccccc}
\toprule
Scale & Models & mAP & NDS & AP@.5& AP@1.& AP@2.& AP@4. & Car & Truck & C.V. & Bus & Trailer & Barrier & Motor. & Bike & Ped. & T.C. \\ 
\midrule
\midrule
\multirow{3}{*}{mini} & BEVFusion\cite{bevfusion-mit} &46.9 &51.5 &44.6 &46.8 &47.8 &48.4 &89.5&	68.5&	0.0&	99.3&	0.0&	0.0&	49.0&	16.8&	91.1&	55.0\\
&\textbf{DGFusion(Ours)} &54.4 &56.7 &52.6 &54.4 &54.9 &55.6 &90.0&70.3&0.0&99.3&0.0&0.0&64.3&50.9&92.6&76.1\\
&\cellcolor{red!20} \textcolor{red}{+} &\cellcolor{red!20} \textit{\textcolor{red}{+7.5}} &\cellcolor{red!20} \textit{\textcolor{red}{+5.2}} &\cellcolor{red!20} \textit{\textcolor{red}{+8.0}} &\cellcolor{red!20} \textit{\textcolor{red}{+7.6}} &\cellcolor{red!20} \textit{\textcolor{red}{+7.1}} &\cellcolor{red!20} \textit{\textcolor{red}{+7.2}} &\cellcolor{red!20} \textit{\textcolor{red}{+0.5}} &\cellcolor{red!20} \textit{\textcolor{red}{+1.8}} & \textit{{+0.0}} &\textit{{+0.0}} &\textit{{+0.0}} &\textit{{+0.0}} &\cellcolor{red!20} \textit{\textcolor{red}{+15.3}} &\cellcolor{red!20} \textit{\textcolor{red}{+34.1}} &
\cellcolor{red!20} \textit{\textcolor{red}{+1.5}} &\cellcolor{red!20} \textit{\textcolor{red}{+21.1}} \\
\midrule
\midrule
\multirow{5}{*}{10\%} & $\ast$CenterPoint\cite{Centerpoint} &47.8 &57.3 &- &- &- &- &79.7&43.7&13.5&59.5&23.3&52.2&46.6&22.4&79.0&57.8\\
& $\ast$GraphAlign++\cite{graphalign++} &53.0 &64.1 &- &- &- &- &82.1&47.4&19.7&63.2&26.8&55.7&55.7&34.2&81.2&63.5\\
& BEVFusion\cite{bevfusion-mit} &62.6 &68.1&49.9&	61.7&67.6&71.2 &86.0&51.2&24.4&73.4&43.3&67.8&68.2&53.8&86.2&71.7\\
&\textbf{DGFusion(Ours)} &64.2 &69.3 &51.5& 62.8& 69.4& 73.0&86.4&61.1&23.3&73.5&42.9&68.4&69.4&55.7&86.5&74.7\\
&\cellcolor{red!20} \textcolor{red}{+} &\cellcolor{red!20} \textit{\textcolor{red}{+1.6}} &\cellcolor{red!20} \textit{\textcolor{red}{+1.2}} &\cellcolor{red!20} \textit{\textcolor{red}{+1.6}} &\cellcolor{red!20} \textit{\textcolor{red}{+1.1}} &\cellcolor{red!20} \textit{\textcolor{red}{+1.8}} &\cellcolor{red!20} \textit{\textcolor{red}{+2.2}} &\cellcolor{red!20} \textit{\textcolor{red}{+0.4}} &\cellcolor{red!20} \textit{\textcolor{red}{+9.9}} & \textit{-1.1} &\cellcolor{red!20} \textit{\textcolor{red}{+0.1}} &\textit{{-0.4}}  &\cellcolor{red!20} \textit{\textcolor{red}{+0.6}} &\cellcolor{red!20} \textit{\textcolor{red}{+1.2}} &\cellcolor{red!20} \textit{\textcolor{red}{+1.9}} &\cellcolor{red!20} \textit{\textcolor{red}{+0.3}} &\cellcolor{red!20} \textit{\textcolor{red}{+3.0}} \\
\midrule
\midrule
\multirow{5}{*}{25\%} & $\ast$CenterPoint\cite{Centerpoint} &56.1 &64.2 &- &- &- &- &83.1&51.1&18.1&61.3&54.6&61.3&54.6&27.3&81.5&68.4\\
& $\ast$GraphAlign++\cite{graphalign++} &62.8 &68.0 &- &- &- &- &86.5&53.5&25.4&64.3&56.7&68.6&68.5&43.7&85.4&75.3\\
&BEVFusion\cite{bevfusion-mit} & 63.3&68.5 &50.2&62.3&68.6&72.2 &86.6&51.0&24.9&72.1&44.4&67.8&70.3&55.7&86.6&73.5\\
&\textbf{DGFusion(Ours)} & 64.6&69.4 & 51.9&63.3&69.9&73.6 &86.6&61.9&25.4&74.5&43.2&68.1&70.4&55.6&86.7&74.2\\
&\cellcolor{red!20} \textcolor{red}{+} &\cellcolor{red!20} \textit{\textcolor{red}{+1.3}} &\cellcolor{red!20} \textit{\textcolor{red}{+0.9}} &\cellcolor{red!20} \textit{\textcolor{red}{+1.7}} &\cellcolor{red!20} \textit{\textcolor{red}{+1.0}} &\cellcolor{red!20} \textit{\textcolor{red}{+1.3}} &\cellcolor{red!20} \textit{\textcolor{red}{+1.4}} &\textit{{+0.0}} &\cellcolor{red!20} \textit{\textcolor{red}{+10.9}} &\cellcolor{red!20} \textit{\textcolor{red}{+0.5}} &\cellcolor{red!20} \textit{\textcolor{red}{+2.4}} &\textit{{-1.2}} &\cellcolor{red!20} \textit{\textcolor{red}{+0.3}} &\cellcolor{red!20} \textit{\textcolor{red}{+0.1}} &\textit{{-0.1}} &\cellcolor{red!20} \textit{\textcolor{red}{+0.1}} &\cellcolor{red!20} \textit{\textcolor{red}{+0.7}} \\
\bottomrule
\end{tabular} 
}
\begin{tablenotes}
\footnotesize
\item[1] [1] `C.V.', `Motor.', `Ped.', and `T.C.' are short for construction vehicle, motorcycle, pedestrian, and traffic cone, respectively.
\item[1] [2] `AP@.5', `AP@1.', `AP@2.', and `AP@4.' represent the mean average precision at distances of 0.5m, 1.0m, 2.0m, and 4.0m from the center, respectively.
\item[1] [3] $\ast$ results are cited from \cite{graphalign++}.
\end{tablenotes}
\label{tab_robustness_samll_scale}
\end{table*}

Reducing the number of samples in the training set and altering the class distribution significantly increases the complexity of inference. To demonstrate the robustness of DGFusion under insufficient training data, we conduct experiments using the nuScenes small-scale dataset for model training and evaluate the performance on both the nuScenes mini validation set and the full validation set. The results are presented in Table~\ref{tab_robustness_samll_scale}. The mini dataset, a subset provided by the official nuScenes team, contains approximately 1.32\% of the full training set, representing an extremely small-scale data environment. The 10\% and 25\% small-scale datasets are constructed following the methodology described in GraphAlign++\cite{graphalign++}.

As shown in Table.~\ref{tab_robustness_samll_scale}, DGFusion outperforms BEVFusion in average performance across all data scales. The reduction in training data converts more objects into hard instances during inference, leading to lower performance of both models compared to that on the full training set (see Table.~\ref{tab_nuscenes_val}). Different experimental setups exhibit distinct trends. First, in the results of the mini version, DGFusion achieves +7.5 mAP and +5.2 NDS, with significant improvements on Motor., Bike, and T.C.. Meanwhile, DGFusion and BEVFusion maintain consistent performance on larger objects like Car, Truck, and Bus. This is attributed to our aggressive EIP/HIP evaluation strategy (see Table.~\ref{tab_ablation_instance_pairs_evaluation}). T.C. achieves improvement through feature matching with Ped., which already has a high AP in the baseline BEVFusion. In contrast, the improvements in Motor. and Bike stem from the stable performance of Car and Truck. Through repeated tests, we find that BEVFusion struggles to obtain stable inference results (mAP $\pm 2.9$) when using the mini version, while DGFusion successfully mitigates this issue (mAP $\pm 0.2$). This benefits from PGIE: when the total amount of data is extremely small, the probability of successful matching by DIPM is low, and PGIE enables stable convergence of the training process through the historical maximum penalty term $\mathcal{L}_{Cos}$. At the 10\% scale, DGFusion achieves +1.6 mAP and +1.2 NDS. At this point, Motorcycle, Bike, and Traffic Cone still show obvious advantages. At the 25\% scale, the performance of the two models becomes closer, but DGFusion still achieves +1.3 mAP and +0.9 NDS. This further indicates that the Dual-guided paradigm can effectively enhance multi-modal BEV features when the data volume and data distribution change.

Second, we notice that the performance of some categories has approached that of the full training process. Among them, the Car of DGFusion has achieved 86.4 AP at the 10\% scale. The Truck shows the most significant improvement, gaining +9.9 AP and +10.9 AP at the 10\% scale and 25\% scale, respectively. This reflects a very important fact that different categories have different requirements for the `training saturation state' \cite{ma2023long_LT3D_2dlatefusion}. This phenomenon requires robust detection to move toward fine-grained enhancement. The strategy of DGFusion to generate L-HIP and C-HIP is a category-based fine-grained enhancement (see Table.~\ref{tab_ablation_instance_pairs_evaluation}).

Finally, through the comparison between the 10\% scale and 25\% scale, it can be observed that a limited increase in data volume does not necessarily lead to improved model performance, such as BEVFusion's Truck (AP 51.2 to 51.0), Bus (AP 73.4 to 72.1), and Barrier (AP 67.8 to 67.8), or DGFusion's Barrier (AP 68.4 to 68.1), Bike (AP 55.7 to 55.6), and Traffic Cone (AP 74.7 to 74.2). We speculate that this phenomenon is related to the limitation of data distribution. The limited data increment at the 25\% scale does not completely change the training balance of the 10\% scale, and even has a negative impact on some categories.

Based on the above discussion, we believe that future work should further consider the balance between two pairs of attributes: one is the total amount of training data and the length of the training process, and the other is the average performance of the model and the performance of individual categories.

\subsubsection{Recall Evaluation}

\begin{figure}[t]
\centering
\includegraphics[width=0.5\textwidth]{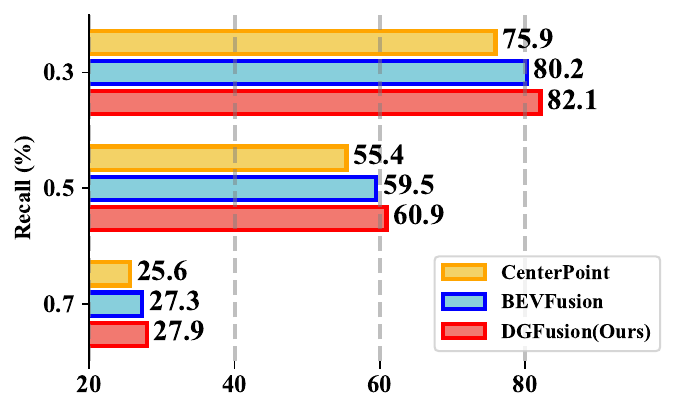}
\caption[ ]{Comparison of average recall on the nuScenes val. set. Different groups represent the calculation of average recall based on different IoU thresholds.} 
\label{fig:recall}
\end{figure}

We compare the Recall\cite{openpcdet,nuscenes} of different models on the nuScenes validation set to further analyze the factors contributing to performance improvement, as shown in Fig.\ref{fig:recall}. Recall is calculated for $IoU=0.3/0.5/0.7$, yielding three sets of results. BEVFusion achieves Recall scores of 80.2/59.5/27.3, depicted in blue. Compared to BEVFusion, DGFusion improves Recall by 1.9/1.4/0.6, represented in red. Since DGFusion employs the CenterPoint head as part of its Instance Match Modules, we also compute the Recall scores of CenterPoint, shown in yellow. The significantly lower Recall of CenterPoint compared to DGFusion suggests that the performance improvement in DGFusion is not primarily due to the use of an additional prediction head for generating instance-level features.



\subsection{Ablation Studys}

\subsubsection{Step-by-step module refinement}

\begin{table}[t]
\scriptsize
\centering
\caption{Step-by-step improvements made by model components.}
\renewcommand\arraystretch{0.95}
\tabcolsep=3.2mm
\resizebox{\linewidth}{!}{
\begin{tabular}{ccccc|cc}
\toprule
\#&E.\&P.&I.&PGIE&IGPE&mAP &NDS \\ 
\midrule
\midrule
(a)&$\surd$& & & &63.3 &68.3 \\
(b)&$\surd$&$\surd$ & & &63.2 &68.0 \\
(c)&$\surd$&$\surd$ &$\surd$ & &63.5 &68.2 \\
(d)&$\surd$&$\surd$ & &$\surd$ &63.8 &68.4 \\
\rowcolor{blue!20} \textbf{(e)}&$\surd$&$\surd$ &$\surd$ &$\surd$ &\textbf{63.9} &\textbf{68.8} \\
\bottomrule
\end{tabular} }
\begin{tablenotes}
\footnotesize
\item[1] [1] All results are from the 25\% scale model. 
\end{tablenotes}
\label{tab_ablation_component}
\end{table}

We conduct an ablation study on the step-by-step improvements made by each module, as shown in Table.\ref{tab_ablation_component}, to validate the effectiveness of the components in the DGFusion pipeline. `E.\&P.' denotes the multi-modal encoders, the final prediction head, and the corresponding loss terms. `I.' represents the Instance Match Modules and its additional prediction loss. The experimental results are from the 25\% scale model. (a) can be seen as a simplified version of BEVFusion, which demonstrates similar performance to the 25\% scale BEVFusion (see Table.\ref{tab_robustness_samll_scale}). The main factors contributing to the performance degradation in (b) are $\mathcal{L}_L$ and $\mathcal{L}_C$; generating instance-level features alone does not affect the model's performance. To separately evaluate PGIE and IGPE, we perform experiments (c) and (d), which can be regarded as the PGI paradigm version and IGP paradigm version of DGFusion, respectively. Compared to (a), (c) improves by 0.2 mAP, and (d) improves by 0.5 mAP. (c) and (d) confirm the positive effects of the Dual-guided Fusion module on BEV features from different modalities, with IGPE demonstrating the best performance. Finally, our strategy in (e) improves by 0.6 mAP and 0.5 NDS compared to (a). 

It is worth noting that the combination of PGIE and IGPE results in a significant improvement in NDS. From the perspective of single-modal hard instance detection, the reliability of C-HIP is generally higher than that of L-HIP (the contribution of features is greater than that of coordinates). Therefore, high-confidence EIP and C-HIP are used in the Camera BEV enhancement process. From a holistic perspective, PGIE and IGPE are symmetric operations, because the Fusion process is essentially a weighted sum of the global BEV, which enables easy camera instance features to act on all BEV spaces. In addition, the performance improvement is independent of the BEV generation method, and all core operations of DGFusion are incremental operations in the unified BEV space. The design of the loss function enables DGFusion to suppress errors in the early stage of training, so as to ensure the detection performance of the original network. DGFusion can be easily extended to any 3D object detection network with a unified space.

\subsubsection{Evaluation strategy for EIP/HIP}

\begin{table}[t]
\scriptsize
\centering
\caption{Impact of evaluation strategies for EIP/HIP.}
\renewcommand\arraystretch{0.95}
\tabcolsep=1.5mm
\resizebox{\linewidth}{!}{
\begin{tabular}{cc|c|cc}
\toprule
\#&Strategy&Scale&mAP&NDS\\ 
\midrule
\midrule
\multirow{2.5}{*}{(a)}&\{Barrier, Ped., T.C.\},&\cellcolor{blue!20} \textbf{mini}&\cellcolor{blue!20} \textbf{52.3}&\cellcolor{blue!20} \textbf{55.8}\\
\cmidrule(lr){3-5}
&\{Car, Truck, C.V., Bus, Trailer, Motor., Bike\}&25\%&63.3 &68.6 \\
\midrule
\midrule
\multirow{2.5}{*}{(b)}&\{Car\},\{Truck, C.V.\},\{Bus, Trailer\},&mini&51.2 &54.7 \\
\cmidrule(lr){3-5}
&\{Barrier\},\{Motor., Bike\},\{Ped., T.C.\}&\cellcolor{blue!20} \textbf{25\%}&\cellcolor{blue!20} \textbf{63.9} &\cellcolor{blue!20} \textbf{68.8} \\
\midrule
\midrule
\multirow{2.5}{*}{(c)}&\multirow{2.5}{*}{None}&mini&50.8 &54.5 \\
\cmidrule(lr){3-5}
&&25\%&62.4 &68.0 \\
\bottomrule
\end{tabular} }
\begin{tablenotes}
\footnotesize
\item[1] [1] `C.V.', `Motor.', `Ped.', and `T.C.' are short for construction vehicle, motorcycle, pedestrian, and traffic cone, respectively.
\end{tablenotes}
\label{tab_ablation_instance_pairs_evaluation}
\end{table}

We observe the evaluation strategies for three types of instance pairs across two datasets of different scales, with the results shown in Table.\ref{tab_ablation_instance_pairs_evaluation}. 
`None' represents that any instance feature pair that successfully matches and passes score filtering will be fused. In the early stages of the experiment, we followed the baseline class grouping strategy proposed by \cite{CBGS}, namely (b), to evaluate EIP and HIP. Strategy (b) demonstrated stronger generalization in the 25\% data scale training process. It is a relatively conservative grouping strategy, considering both the total number of targets in each class and the shape and size variations between different classes. However, it is worth noting that in the mini-scale experiment, strategy (b) struggled to consistently generate stable results, a phenomenon similar to BEVFusion's performance at the mini scale, despite the superior performance of DGFusion in this context. This observation highlights a notable drawback of strategy (b) when the data scale changes, which we speculate may be related to significant variations in class similarity during the training process. 

Inspired by the use of a hierarchical structure to categorize all classes in \cite{peri2023towardsLT3D}, we aim to define class similarity using a criterion independent of data distribution. Therefore, we propose strategy (a) to mitigate this issue. Compared to (b), (a) is a more aggressive grouping strategy. We aim to decide whether feature interaction between classes occurs based on the collision cost when the dataset is small. The collision cost is derived from real-world driving experiences and has relatively low correlation with the data distribution in 3D object detection tasks. From a safety perspective in autonomous driving, the cost of collisions with barriers, pedestrians, and traffic cones is particularly high, as barriers and traffic cones typically mark boundaries between different traffic zones. Feature interaction evaluation based on collision cost will be one of the optimization directions in future work. Strategy (c) does not remove any instance pairs, regardless of the scale. The results of (c) confirm the necessity of evaluating and filtering instance pairs.

\subsubsection{Latency analysis for models/components}

\begin{table}[t]
\scriptsize
\centering
\caption{Latency analysis for models/components.}
\renewcommand\arraystretch{0.9}
\tabcolsep=6mm
\resizebox{\linewidth}{!}{
\begin{tabular}{lc}
\toprule
Models/Components &Latency(ms) \\ 
\midrule
\midrule
TransFusion-L\cite{bai2022transfusion}&82.3\\
\midrule
FocalFormer3D-L\cite{chen2023focalformer3d}&715.4\\
\midrule
BEVFusion\cite{bevfusion-mit}&139.6\\
\midrule
\rowcolor{blue!20} \textbf{DGFusion(Ours)}& \textbf{219.9}\\
- Multi-modal Encoders& 84.7\\
- Instance Match Modules-IFG& 83.1\\
- Instance Match Modules-DIPM& 0.2\\
- Dual-guided Modules& 1.8\\
- Prediction(BEV Backbone+Head)& 50.1\\
\bottomrule
\end{tabular} }
\label{tab_ablation_latency}
\end{table}
We conduct a latency analysis of DGFusion, as shown in Table.\ref{tab_ablation_latency}. The models used for comparison include the single-modal models TransFusion-L and FocalFormer3D-L, as well as the multi-modal model BEVFusion. To ensure fairness, all models were retested on the workstation mentioned in Sec.\ref{subsec:Details}. TransFusion-L was trained and tested on OpenPCDet\cite{openpcdet}. FocalFormer3D-L trained and tested on MMDetection3D\cite{mmdet3d2020}. The time consumption of DGFusion's Multi-modal Encoders and prediction are comparable to that of BEVFusion (134.8ms vs. 139.6ms). IFG requires 83.1ms, while DIPM only takes 0.2ms. The majority of the time consumption in the Instance Match Modules comes from IFG, indicating potential optimization opportunities in future work. The Dual-guided Modules requires only 1.8ms. Although FocalFormer3D demonstrates superior performance, it comes with a significantly higher time cost, approximately 3 times greater than that of DGFusion.

\subsubsection{Sampling strategy for instance-level features}

\begin{table}[t]
\scriptsize
\centering
\caption{Impact of different sampling strategies on instance-level features.}
\renewcommand\arraystretch{0.95}
\tabcolsep=3.8mm
\resizebox{\linewidth}{!}{
\begin{tabular}{cccc|cc}
\toprule
\#&Center&Vertex&Boundary-Mid&mAP&NDS\\ 
\midrule
(a)&$\surd$& & &63.1 &68.5 \\
(b)&$\surd$&$\surd$& &63.2 &68.3 \\
\rowcolor{blue!20} \textbf{(c)}&$\surd$& &$\surd$&\textbf{63.9} &\textbf{68.8} \\
(d)&$\surd$&$\surd$&$\surd$&63.9 &68.6 \\
\bottomrule
\end{tabular} }
\begin{tablenotes}
\footnotesize
\item[1] [1] All results are from the 25\% scale model. 
\end{tablenotes}
\label{tab_ablation_instance_feature_type}
\end{table}

We investigate the impact of different sampling strategies for instance-level features on the performance of DGFusion, as shown in Table.\ref{tab_ablation_instance_feature_type}. `Center' refers to the feature of the proposal's center point, `Vertex' refers to the features of the four vertices of the proposal, and `Boundary-Mid' refers to the features of the midpoints of the four boundaries of the proposal. All strategies follow the feature concatenation order as described in the table. The `Center' feature serves as the core of the instance-level features. (c) represents the strategy we have used, which has the same meaning as Fig.\ref{fig:CameraBEV_Enhancement_and_Instance_Generation}.(b). (d) adds the four vertex features to (c), achieving similar performance. However, (b) suggests that relying solely on the four vertex features to describe the candidate region is insufficient, likely due to the relatively limited valid information contained in the vertices of the candidate region.

\subsubsection{Score filtering threshold $\gamma$}

\begin{table}[t]
\scriptsize
\centering
\caption{Impact of $\gamma$ in IFG.}
\renewcommand\arraystretch{0.95}
\tabcolsep=3.7mm
\resizebox{0.5\linewidth}{!}{
\begin{tabular}{c|cc}
\toprule
$\gamma$&mAP&NDS\\ 
\midrule
\midrule
0.5&63.7&68.3\\
\rowcolor{blue!20} \textbf{0.7}&\textbf{63.9}&\textbf{68.8}\\
0.9&63.8&68.9\\
\bottomrule
\end{tabular} }
\label{tab_ablation_gamma}
\begin{tablenotes}
\footnotesize
\item[1] [1] All results are from the 25\% scale model. 
\end{tablenotes}
\end{table}

We investigate the impact of the threshold $\gamma$ for score filtering in IFG on the performance of DGFusion, as shown in Table.\ref{tab_ablation_gamma}. Experimental results indicate that changes in $\gamma$ have a minimal effect on mAP. The best performance is achieved when $\gamma=0.7$. When $\gamma=0.5$, a noticeable decline in NDS occurs. During the early stages of training, we observed a clear correlation between $\gamma$ and the occurrence of empty EIP and HIP: the larger the value of $\gamma$, the higher the probability of EIP and HIP being empty. This aligns with our design intent, as the additional detection head does not load any pre-trained parameters, and the proposals generated during the early stages of training are of low quality. A larger $\gamma$ helps suppress errors from the additional detection head. In a sufficiently long training process, using a larger $\gamma$ is more reasonable.

\section{Conclusion}


In this work, we address the challenging problem of hard instance detection in autonomous driving by analyzing the information density gap across modalities. Our proposed DGFusion framework introduces a dual-guided paradigm that synergizes the complementary strengths of Point-guided-Image and Image-guided-Point approaches. Specifically, the Difficulty-aware Instance Pair Matcher (DIPM) dynamically identifies hard instances, while the Dual-guided Modules adaptively fuses multi-modal features to enhance BEV representations. Extensive experiments on nuScenes show that DGFusion achieves consistent improvements over baseline methods, particularly for distant, occluded, and small objects.

\textbf{Limitation and Future Work.}
A key limitation of DGFusion is that the quality of instance-level features restricts the model's performance, and their generation process introduces the most significant additional time cost. Our future work will focus on enhancing instance representations while maintaining the established Dual-guided paradigm.

\bibliographystyle{IEEEtran}
\bibliography{review}

\begin{IEEEbiography}
[{\includegraphics[width=1in,height=1.25in,clip,keepaspectratio]{{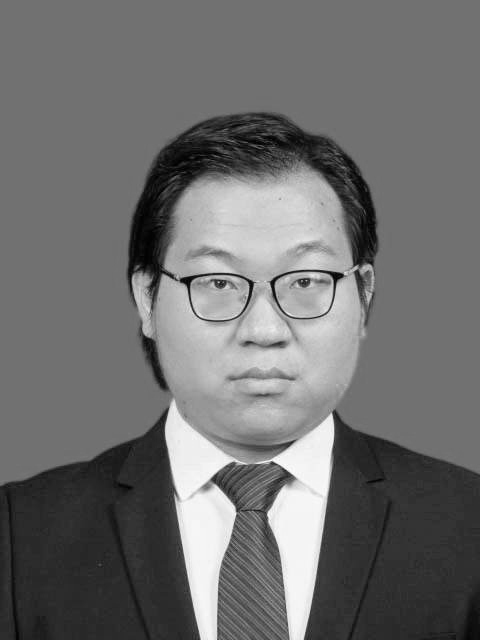}}}]
{Feiyang Jia} was born in Yinchuan, Ningxia Province, China, in 1998. He received his B.S. degree from Beijing Jiaotong University (China) in 2020. He received a master's degree from Beijing Technology and Business University (China) in 2023. He is now a Ph.D. student majoring in Computer Science and Technology at Beijing Jiaotong University (China), with research focus on Computer Vision.
\end{IEEEbiography}

\begin{IEEEbiography}[{\includegraphics[width=1in,height=1.25in,clip,keepaspectratio]{{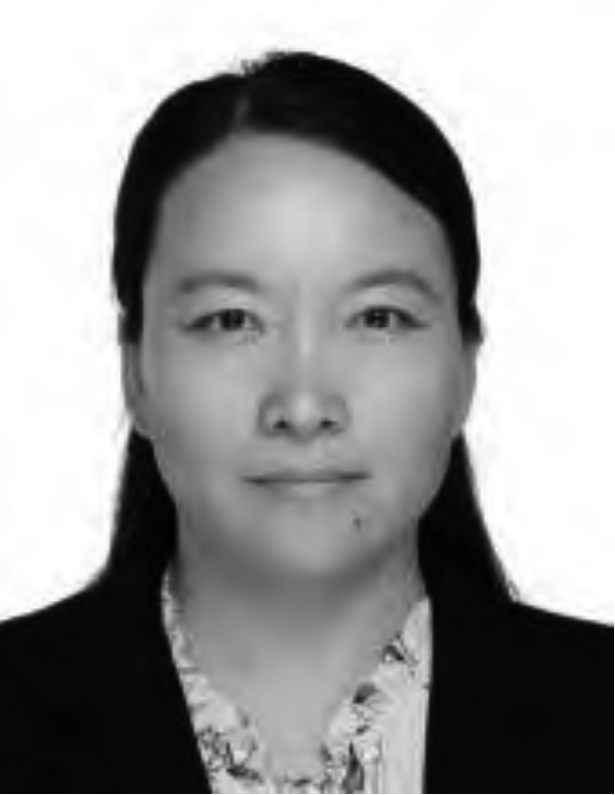}}}]{Caiyan Jia} was born in 1976. She received her Ph.D. degree from Institute of Computing Technology, Chinese Academy of Sciences, China, in 2004. She had been a postdoctor in Shanghai Key Lab of Intelligent Information Processing, Fudan University, Shanghai, China, in 2004–2007. She is now a professor in School of Computer and Information Technology, Beijing Jiaotong University, Beijing, China. Her current research interests include deep learning in computer vision, graph neural networks and social computing, etc.
\end{IEEEbiography}

\begin{IEEEbiography}
[{\includegraphics[width=1in,height=1.25in,clip,keepaspectratio]{{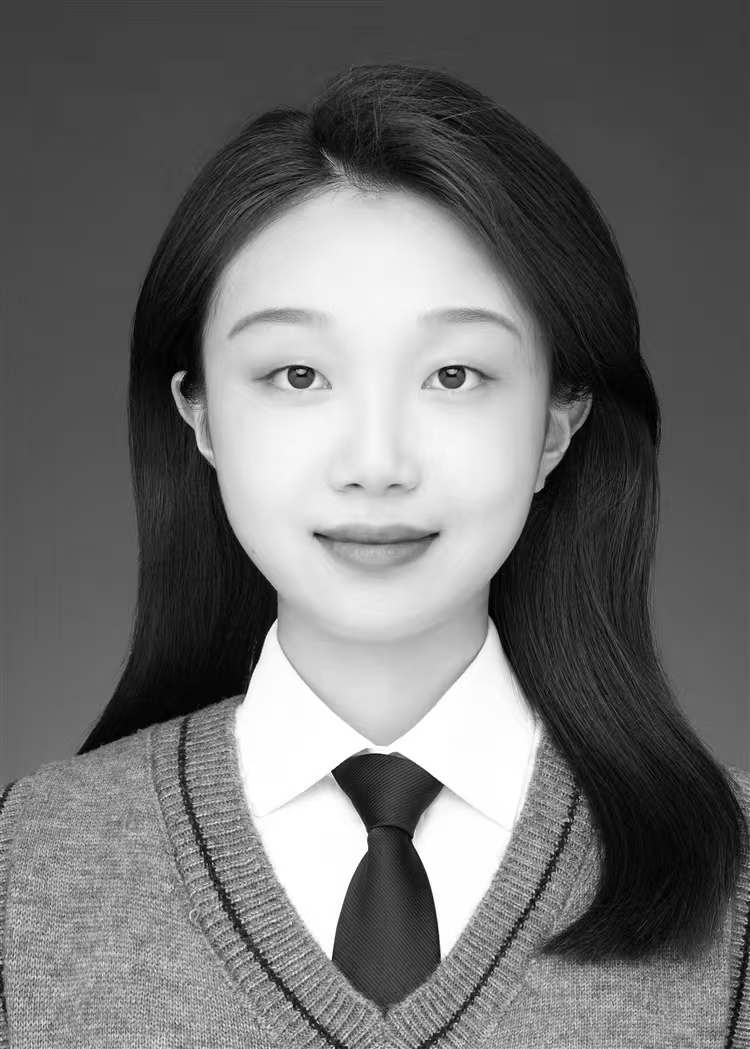}}}]{Ailin Liu} was born in Dalian, Liaoning Province, China, in 2002. She received her B.S. degree from Beijing Forestry University (China) in 2024. She is now a master student majoring in Computer Science and Technology at Beijing Jiaotong University (China), with research focus on Computer Vision.
\end{IEEEbiography}

\begin{IEEEbiography}[{\includegraphics[width=1in,height=1.25in,clip,keepaspectratio]{{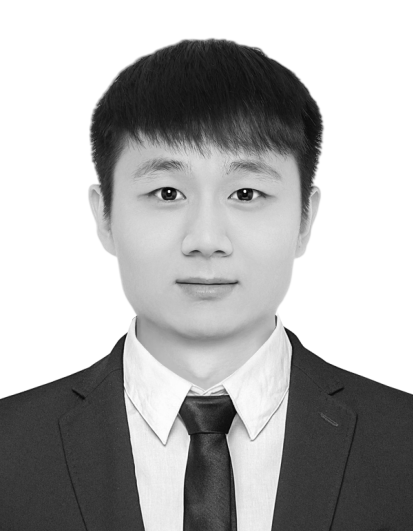}}}]{Shaoqing Xu} received his M.S. degree in transportation engineering from the School of Transportation Science
and Engineering in Beihang University. He is currently working toward the Ph.D. degree in electromechanical engineering with the State Key Laboratory of Internet of Things for Smart City, University of Macau, Macao SAR, China. His research interests include intelligent transportation systems, Robotics and computer vision.
\end{IEEEbiography}

\begin{IEEEbiography}[{\includegraphics[width=1in,height=1.25in,clip,keepaspectratio]{{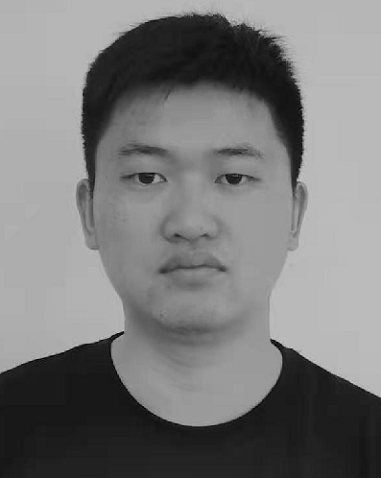}}}]{Qiming Xia} received the M.S. degree from Jimei University, Xiamen, China, in 2022. He is currently working toward a Ph.D. degree in the School of Informatics at Xiamen University, Xiamen, China. His research interests include computer vision, machine learning, and 3D object detection. 
\end{IEEEbiography}

\begin{IEEEbiography}
[{\includegraphics[width=1in,height=1.25in,clip,keepaspectratio]{{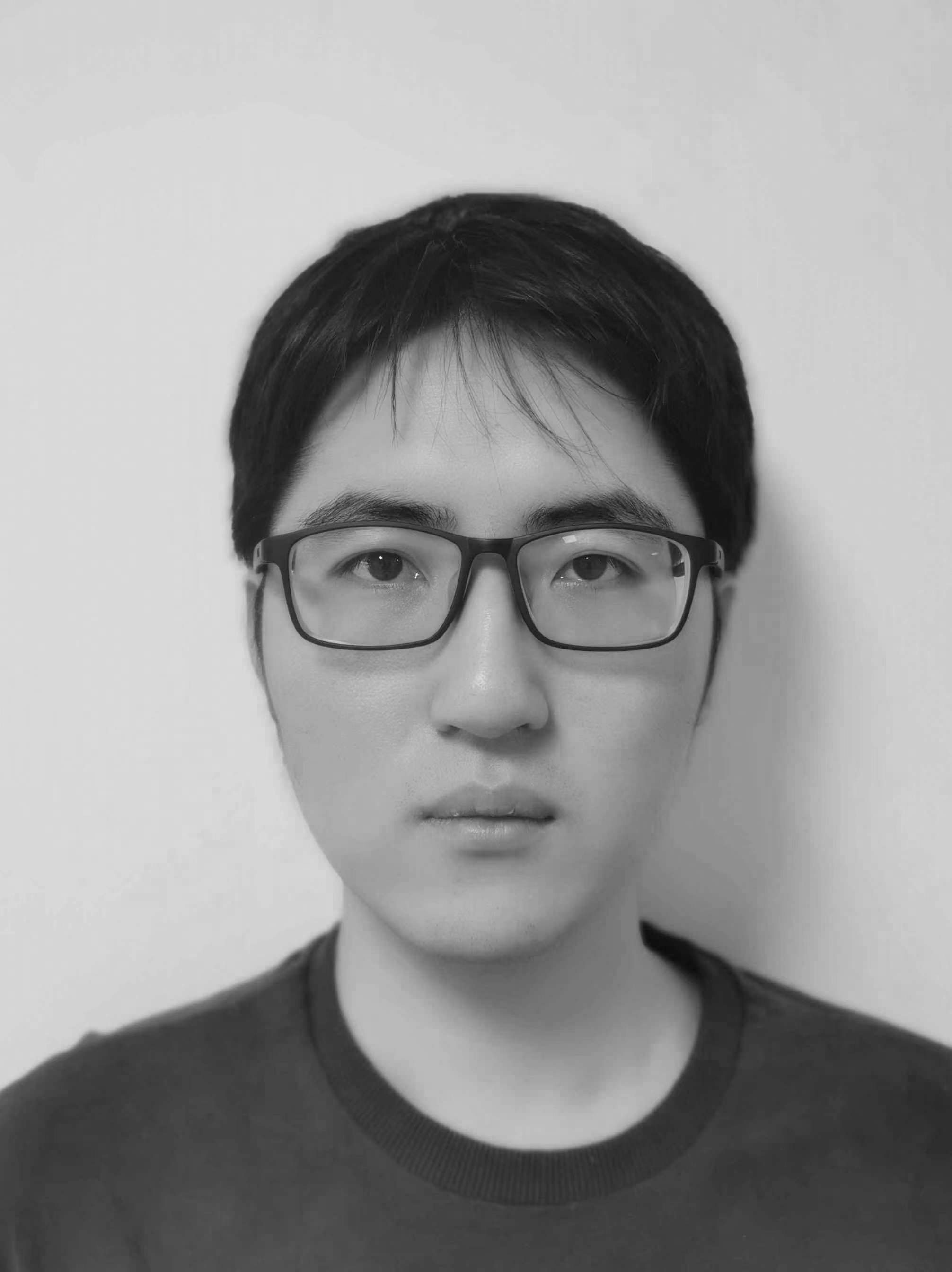}}}] {Lin Liu} was born in Jinzhou, Liaoning Province, China, in 2001. He is now a college student majoring in Computer Science and Technology at China University of Geosciences(Beijing). Since Dec. 2022, he has been recommended for a master's degree in Computer Science and Technology at Beijing Jiaotong University. His research interests are in computer vision.
\end{IEEEbiography}


\begin{IEEEbiography}[{\includegraphics[width=1in,height=1.25in,clip,keepaspectratio]{{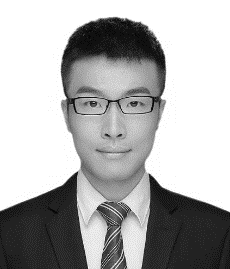}}}]
{Lei Yang (Member, IEEE)} received his M.S. degree from the Robotics Institute at Beihang University, in 2018. and the Ph.D. degree from the School of Vehicle and Mobility, Tsinghua University, in 2024. From 2018 to 2020, he joined the Autonomous Driving R\&D Department of JD.COM as an algorithm researcher. Currently, he is a research fellow with the School of Mechanical and Aerospace Engineering, Nanyang Technological University, Singapore. His current research interests include autonomous driving, 3D scene understanding and world model.
\end{IEEEbiography}

\begin{IEEEbiography}[{\includegraphics[width=1in,height=1.25in,clip,keepaspectratio]{{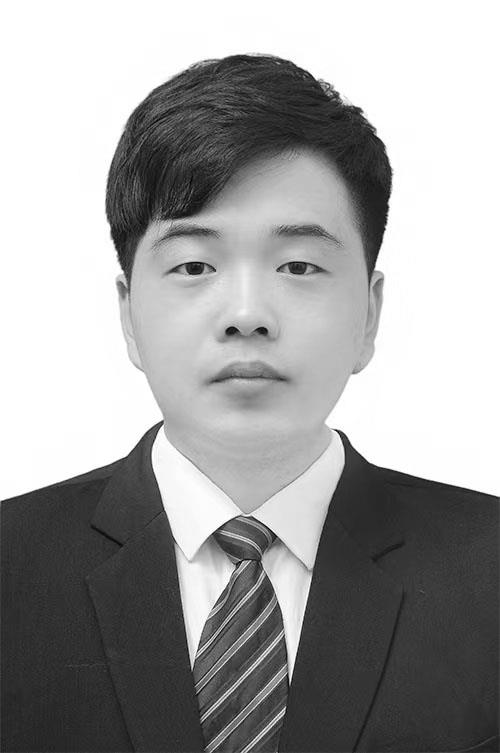}}}]{Yan Gong} received a M.S. degree in Computer Sci- ence and Technology from Northeastern University, Shenyang, China, in 2023. He served as a research assistant with the School of Vehicle and Mobility at Tsinghua University, Beijing, China. He is now a researcher at Autonomous Driving Division of X Research Department, JD Logistics. He has authored and co-authored many academic papers. His research interests include pattern recognition, multi-modal fusion and deep learning.

\end{IEEEbiography}
\begin{IEEEbiography}[{\includegraphics[width=1in,height=1.25in,clip,keepaspectratio]{{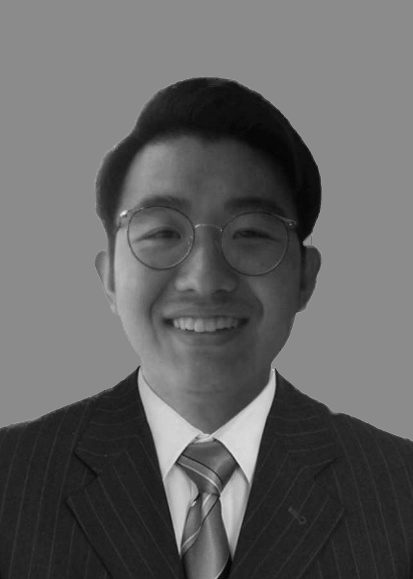}}}]{Ziying Song} was born in Xingtai, Hebei Province, China, in 1997. He received his B.S. degree from Hebei Normal University of Science and Technology (China) in 2019. He received a master's degree from Hebei University of Science and Technology (China) in 2022. He is now a Ph.D. student majoring in Computer Science and Technology at Beijing Jiaotong University (China), with research focus on Computer Vision. 
\end{IEEEbiography}

\end{document}